\documentclass[10pt,twocolumn,letterpaper]{article}

\usepackage{cvpr}
\usepackage{times}
\usepackage{epsfig}
\usepackage{graphicx}
\usepackage{amsmath}
\usepackage{amssymb}
\usepackage{float}
\usepackage{booktabs} 
\usepackage{textcomp}
\usepackage{multirow}
\usepackage{amssymb}
\usepackage{makecell}
\usepackage{notoccite}
\usepackage{comment}
\usepackage{subfiles}
\newcommand{\red}[1]{{\color{red}#1}}
\renewcommand{\etal}{\textit{et al.~}}
\def\eg{\textit{e.g.},~}
\def\ie{\textit{i.e.},~}

\usepackage[pagebackref=true,breaklinks=true,letterpaper=true,colorlinks,bookmarks=false]{hyperref}

\usepackage[breaklinks=true,bookmarks=false]{hyperref}

\cvprfinalcopy  
\def\cvprPaperID{4442}  
\begin{document}

\title{Learning to Optimize Non-Rigid Tracking}
\author{
Yang Li${}^\text{1,4}$ Alja{\v{z}} Bo{\v{z}}i{\v{c}}${}^\text{4}$ Tianwei Zhang${}^\text{1}$ Yanli Ji${}^\text{1,2}$ Tatsuya Harada${}^\text{1,3}$ Matthias Nie{\ss}ner${}^\text{4}$\\
\\
${}^\text{1}$The University of Tokyo, ${}^\text{2}$UESTC, ${}^\text{3}$RIKEN, ${}^\text{4}$Technical University Munich\\}
\maketitle

\begin{abstract}
One of the widespread solutions for non-rigid tracking has a nested-loop structure: with Gauss-Newton to minimize a tracking objective in the outer loop, and Preconditioned Conjugate Gradient (PCG) to solve a sparse linear system in the inner loop.
In this paper, we employ learnable optimizations to improve tracking robustness and speed up solver convergence.
First, we upgrade the tracking objective by integrating an alignment data term on deep features which are learned end-to-end through CNN.  The new tracking objective can capture the global deformation which helps Gauss-Newton to jump over local minimum, leading to robust tracking on large non-rigid motions. Second, we bridge the gap between the preconditioning technique and learning method by introducing a ConditionNet which is trained to generate a preconditioner such that PCG can converge within a small number of steps. Experimental results indicate that the proposed learning method converges faster than the original PCG by a large margin. 
\end{abstract}

\section{Introduction}
Non-rigid dynamic objects, \eg humans and animals, are important targets in both computer vision and robotics applications. Their complex geometric shapes and non-rigid surface changes result in challenging problems for tracking and reconstruction. 
In recent years, using commodity RGB-D cameras,
the seminal works such as DynamicFusion \cite{dynamicfusion} and VolumeDeform \cite{volumedeform}
made their efforts to tackle this problem and obtained impressive non-rigid reconstruction results. 
At the core of DynamicFucion and VolumeDeform are non-linear optimization problems. However, this optimization can be slow, and can also result in undesired local minima. In this paper, we propose a learning-based method that finds optimization steps that expand the convergence radius (\ie avoids local minima) and also makes convergence faster. 
We test our method on the essential inter-frame non-rigid tracking task, \ie to find the deformation between two RGB-D frames, which is a high-dimensional and non-convex problem. 
The absence of an object template model, large non-overlapping area, and observation noise in both source and target frame make this problem even more challenging. 
This section will first review the classic approach and then put our contributions into context.

\begin{figure}[t!]
    \centering
    \includegraphics[width=1\linewidth]{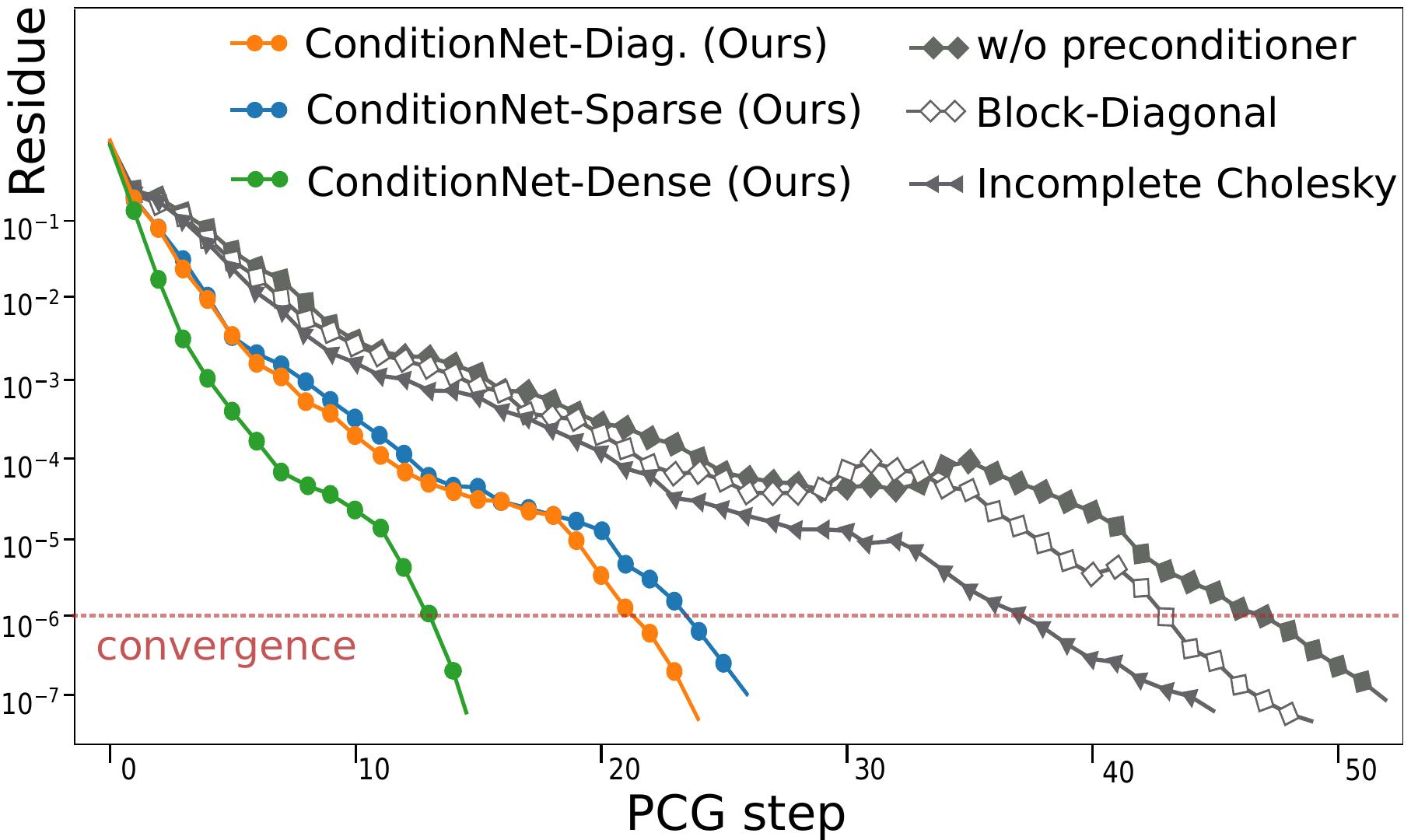}
    \caption{PCG convergence using different preconditioners. The curves show the average convergence on the testing dataset. Note that our final method (green curve) requires 3 times fewer PCG steps to achieve the same residual ($10^{-6}$) than the best baseline (dashed line).}
    \label{fig:pcg_convergence}
\end{figure}

\paragraph{Non-rigid Registration} 
The non-rigid surface motions can be roughly approximated through the ``deformation graph" \cite{embededdeformation}. In this deformable model, all of the unknowns, \ie the rotations and translations, are denoted as $\mathcal{G}$. 
Given two RGB-D frames, the goal of non-rigid registration is to determine the $\mathcal{G}$ that minimizes the typical objective function:
\begin{equation}
\label{eqn:total_energy}
\min_{\mathcal{G} } \{\mathbf{E}_{fit}(\mathcal{G})   + \lambda\mathbf{E}_{reg}(\mathcal{G}) \}
\end{equation}
where $\mathbf{E}_{fit}$ is the data fitting term that measures the closeness between the warped source frame and the target frame. 
Many different data fitting terms have been proposed over the past decades, such as the geometric point-to-point and point-to-plane constraints~\cite{li2008regist,zollhofer2014real,dynamicfusion,volumedeform}
sparse SIFT descriptor correspondences \cite{volumedeform}, and the dense color term \cite{zollhofer2014real}, \textit{etc}. 
The term $\mathbf{E}_{reg}$ regularizes the problem by favoring locally rigid deformation. Coefficient $\lambda$ balances these two terms.
The energy (\ref{eqn:total_energy}) is minimized by iterating the Gauss-Newton update step \cite{Gauss_newton} till convergence.
Inside each Gauss-Newton update step a large linear system needs to be solved, for which an iterative preconditioned conjugate gradient (PCG) solver is commonly used.

This classic approach cannot properly handle large non-rigid motions since the data fitting term $\mathbf{E}_{fit}$ in the energy function (\ref{eqn:total_energy}) is made of local constraints (\eg dense geometry or color maps), which only work when they are close to the global solution, or global constraints that are prone to noise (\eg sparse descriptor). In the case of large non-rigid motions, these constraints cannot provide convergent residuals and lead to tracking failure.
In this paper, we alleviate the non-convexity of this problem by introducing a deep feature alignment term into $\mathbf{E}_{fit}$. The deep features are extracted through an end-to-end trained CNN. We assume that, by leveraging the large receptive field of convolutional kernels and the nature of the data-driven method, the learned feature can capture the global information which helps Gauss-Newton to jump over local minimums.

\begin{figure}[htb!]
	\centering
	\includegraphics[width=0.9\linewidth]{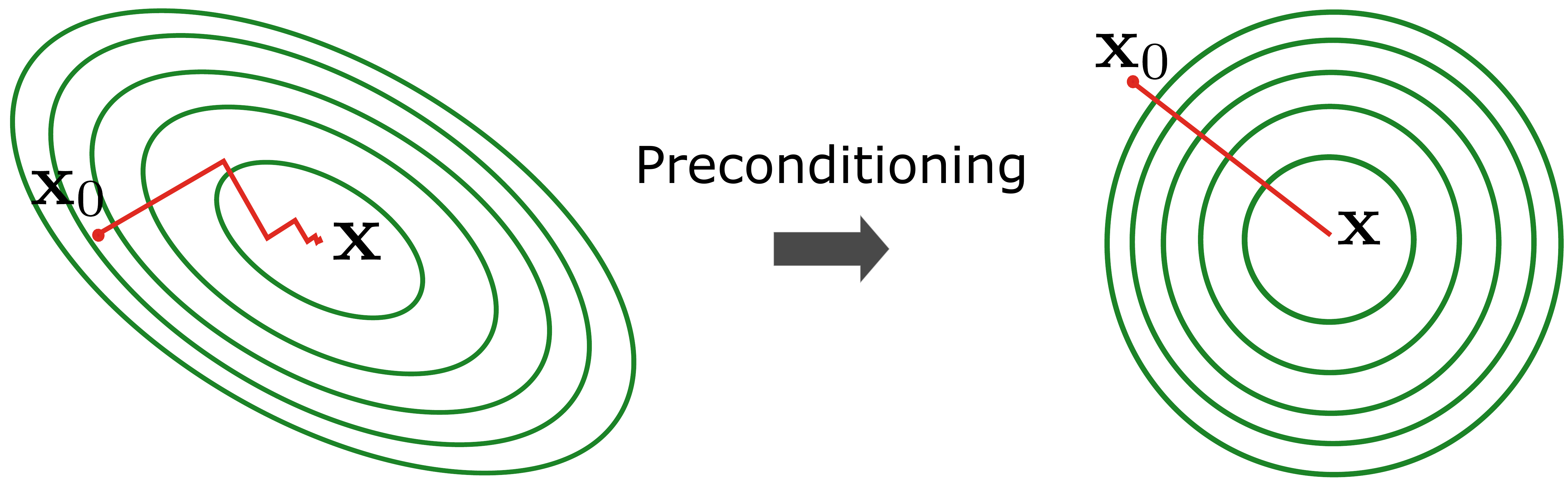}      
	\caption{Example of using the Deepest Decent to solve a 2D system. Deepest Decent needs multiple steps to converge on an ill-conditioned system (left) and only one step on a perfectly conditioned system (right). Intuitively, Preconditioning is trying to modify the energy landscape from an elliptical paraboloid into a spherical one such that from any initial position, the direction of the first-order derivative directly points to the solution.}
	\label{fig:preconditioning}
\end{figure}

As illustrated in Fig. \ref{fig:preconditioning}, preconditioning speeds up the convergence of an iterative solver. The general idea behind preconditioning is to use a matrix, called \textit{preconditioner}, to modify an ill-posed system into a well-posed one that is easier to solve. 
As the hard-coded block-diagonal preconditioner was not designed specifically for the non-rigid tracking task, the existing non-rigid tracking solvers are still time-consuming. We argue that PCG converges  much faster if the design of the preconditioner involves prior expert knowledge of this specific task.
Then we raise the question: \textit{does the data-driven method learn a good preconditioner?} In this paper, we exploit this idea by training neural network to generate a preconditioner such that PCG can converge within a few steps.

Our contribution is twofold:
\begin{itemize}
    
    \item We introduce a deep feature fitting term based on end-to-end learned CNN for the non-rigid tracking problem.
    Using the proposed data fitting term, the non-rigid tracking Gauss-Newton solver can converge to the global solution even with large non-rigid motions.
    \item We propose ConditionNet that learns to generate a problem-specific preconditioner using a large number of training samples from the Gauss-Newton update equation. The learned preconditioner increases PCG's convergence speed by a large margin.
\end{itemize}

\section{Related Works}
\subsection {Classic Data-terms for Non-rigid Tracking}
The core of non-rigid tracking is to define a data fitting term for robust registration.
Many different data fitting terms have been proposed in the recent geometric approaches, \eg the point-to-point alignment terms in \cite{li2008regist}, and the point-to-plane alignment terms in \cite{dynamicfusion,volumedeform}.
Beside dense geometric constraints, sparse color image descriptor detection and matching have been used to establish the correspondences in \cite{volumedeform}.
In additions, in \cite{zollhofer2014real}, the potential of color consistency assumption was studied. 
Furthermore, to deal with the lighting change, the reflection consistence technique was proposed in \cite{guo2017real}, and the correspondence prediction using decision trees was developed in \cite{dou2016fusion4d}.

\subsection{Learning based tracking}
This line of research focuses on solving motion tracking tasks from a deep learning perspective. 
One of the promising ideas is to replace the hand-engineered descriptors with the learned ones. 
For instance, the Learned Invariant Feature Transform (LIFT) is proposed in \cite{lift}, the volumetric descriptor for 3D matching is proposed in \cite{3dmatch}, and the coplanarity descriptor for plane matching is proposed in~\cite{planematch}. 
For non-rigid localization/tracking, Schmidt~\etal~\cite{schmidt2016self} use Fully-Convolutional networks to learn the dense descriptors for upper torso and head of the same person;  
Alja{\v{z}}~\etal~\cite{deepdeform} proposed a large labeled dataset of sparse correspondence for general non-rigidly deforming objects, and a Siamese network based non-rigid 3D patch matching approach. 
Regression networks have also been used to directly map input sensor data to motions, including the camera pose tracking~\cite{deeptam}, the dense optical flow tracking~\cite{flownet}, and the 3D scene flow estimation \cite{flownet3d}. 
The problem of motion regression is that the regressors could be overwhelmed by the complexity of the task, therefore, leading to severe over-fitting. 
A more elegant way is to let the model focus on a simple task, such as feature extraction while using classic optimization tools to solve the rest. This resulted in the recent works that combine Gauss-Newton optimization and deep learning to learn the most suitable features for image alignment~\cite{cascade_lucas_kanade}, pose registration~\cite{regnet,deepicn}, and multi-frame direct bundle-adjustment~\cite{banet}. 
Inspired by these works, we integrate the entire non-rigid optimization method into the end-to-end pipeline to learn the optimal feature for non-rigid tracking, which requires dealing with orders of magnitude more degree of freedoms than the previous cases. 
The details are described in Section \ref{section:feature_learning}.

\subsection{Preconditioning Techniques} 
Preconditioning as a method of transforming a difficult problem into one that is easier to solve has centuries of history. Back to 1845, the Jacobi's Method~\cite{jacobi_method} was first proposed to improve the convergence of iterative methods. Block-Jacobi is the simplest form of preconditioning, in which the preconditioner is chosen to be the block diagonal of the linear system that we want to solve. Despite its easy accessibility, we found that applying it shows only a marginal improvement in our problem. Other methods, such as Incomplete Cholesky Factorization, multiGrid method~\cite{tatebe1993multigrid} or successive over-relaxation~\cite{van1989high} method have shown their effectiveness in many applications. In this paper, we exploit the potential of data-driven preconditioner to solve the linear system in the non-rigid tracking task. The details are shown in Section \ref{section:conditioner}.

\section{Learning Deep Non-Rigid Feature}\label{section:feature_learning}

\subsection{Scene Representation}

The input of our method is two frames that are captured using a commodity RGB-D sensor. Each frame contains a color map and a depth map both at the size of  640$\times$480. Calibration was done to ensure that color and depth were aligned in temporal and spatial domain. We denote the source frame as $\mathbf{S}$, and the target frame as $\mathbf{T}$. 

\begin{figure}[t!]
    \centering
    \includegraphics[width=1\linewidth]{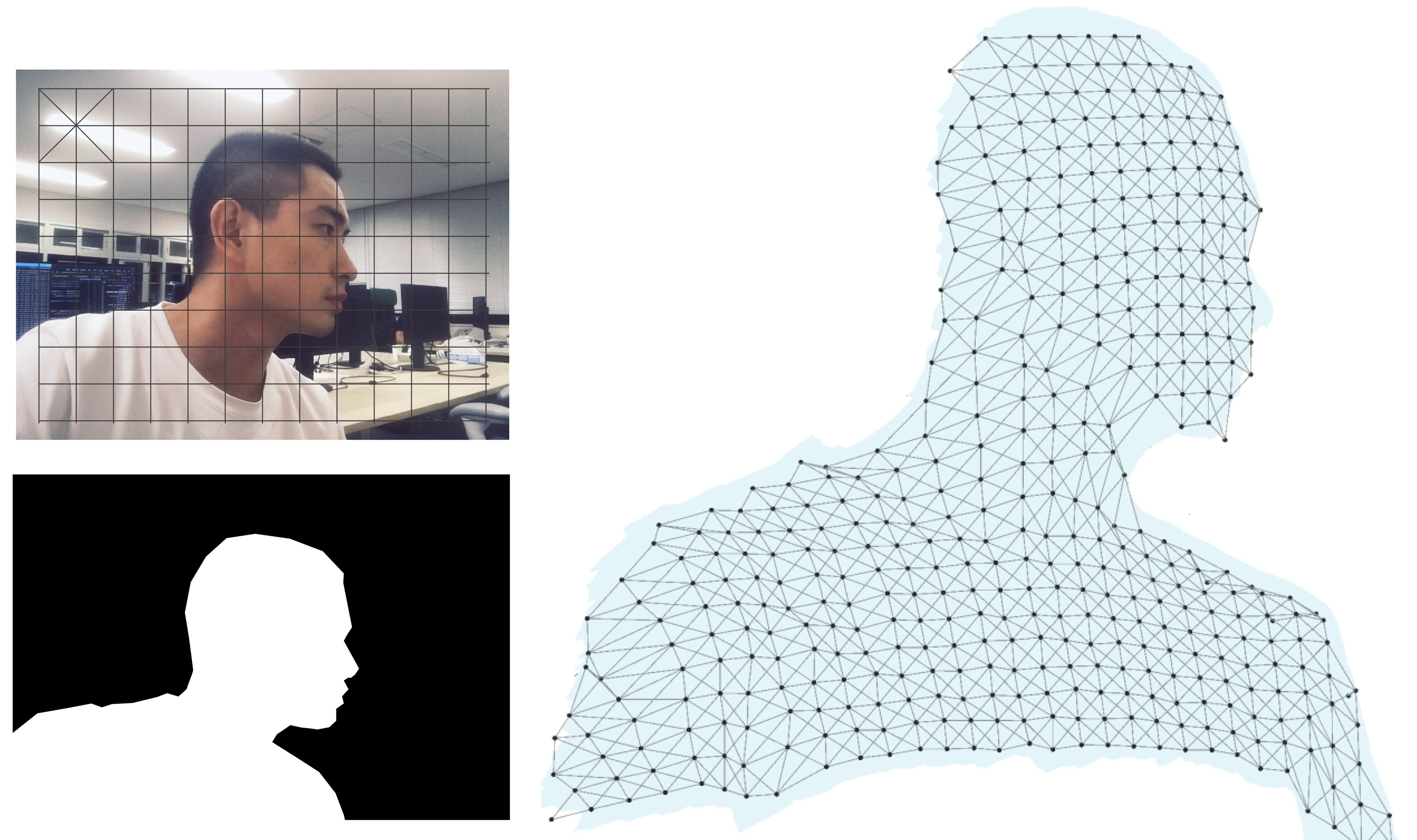} 
    \caption{Our deformation graph. Left top: Uniform sampling on the pixel grid. Let bottom: Binary mask acquired using simple depth threshold or depth aided human annotation. Right:  Masked 3D deformation graph.
    }
    \label{fig:trinity}
    \centering

\end{figure}

We approximate the surface deformation with the deformation graph $\mathcal{G}$. Fig. \ref{fig:trinity} shows an example of our deformation graph.
We uniformly sample the image, resulting in a rectangle mesh grid of size $ w \times h$. 
A point in the mesh grid is treated as a node in the deformation graph. Each node connects exactly to its 8 neighboring nodes. To filter out the invalid nodes, a binary mask $\mathcal{V} \in \mathbb{R}^{w\times h}$ is constructed by checking if the node is from the background, holds invalid depth, or lies on occlusion boundaries with large depth discontinuity. Similarly, edges are filtered by the mask $\mathcal{E} \in \mathbb{R}^{w\times h\times 8}$ if they link to invalid nodes or go beyond the edge length threshold.  In the deformation graph, the node $i$ is parameterized by a translation vector $\mathbf{t}_i \in \mathbb{R}^3 $ and a matrix $ \mathbf{R}_i \in \mathrm{SO}3 $. Putting all parameters into a single vector, we get
$$\mathcal{G}=\{ \mathbf{R}_i, \mathbf{t}_i |_{ i = 1, 2,  \cdots, w \times h} \} $$

\subsection{Deep Feature Fitting Term}
We use the function $\mathcal{F}(\cdot)$, which is based on fully convolutional networks~\cite{fcn}, to extract feature map from source frame $\mathbf{S}$ and target frame $\mathbf{T}$. The encoded feature maps are:
\begin{equation}
\mathbf{F_S} = \mathcal{F}(\mathbf{S}), \;\;\;\;    \mathbf{F_T} = \mathcal{F}(\mathbf{T})
\end{equation}

We apply up-sampling layers in the neural network such that the encoded feature map has the size $w\times h\times c$, where $c$ is the dimension of a single feature vector. Thus the feature map and the deformation graph have the same rows and columns.  This means that a feature vector and a graph node have a one-to-one correspondence (to reduce GPU memory overhead and speed up the learning).  
We denote $\mathbf{D_S} \in \mathbb{R}^{w\times h}$ and $\mathbf{D_T} \in \mathbb{R}^{w\times h}$ as the sampled depth map from source and target frames. Given the translation vectors $\mathbf{t}_i \in \mathcal{G}$, and the depth value $\mathbf{D_T}(i)$, 
the projected feature for the pixel $i$ can be obtained by
\begin{equation}
\mathbf{\tilde{F}_S}(i) = \mathbf{F_S} ( \pi ( \mathbf{t}_i, \mathbf{D_T}(i)) )
\end{equation}
where $\pi(\cdot):\mathbb{R}^2 \rightarrow \mathbb{R}^2 $ is the warping function that maps one pixel coordinate to another pixel coordinate by applying translation $\mathbf{t}_i$ to a back-projected pixel $i$, and projecting the transformed point to the source camera frame.  The warped coordinate are continuous values. $\mathbf{\tilde{F}_S}(i)$ is sampled by bi-linearly interpolating the 4 nearest features on the 2D mesh grid. 
This sampling operation is made differentiable using the spatial transformer network defined in \cite{spatial_transformer}. Then the \textbf{deep feature fitting term} is defined as
\begin{equation}
\label{eqn:feature_term}
\mathbf{E}_{fea}(\mathcal{G}) = \lambda_{f} \sum^{w \times h}_{i=0}  \mathcal{V}_i \cdot||  \mathbf{
    \tilde{F}_S}(i)- \mathbf{F_T}(i)  ||^2
\end{equation} 
Note that compared to the classic color-consistency constraints, the learned deep feature captures high-order spatial deformations in the scans, by leveraging the large receptive field size of the convolution kernels.

\begin{figure*}[ht]
    \centering
    \includegraphics[width=1\linewidth]{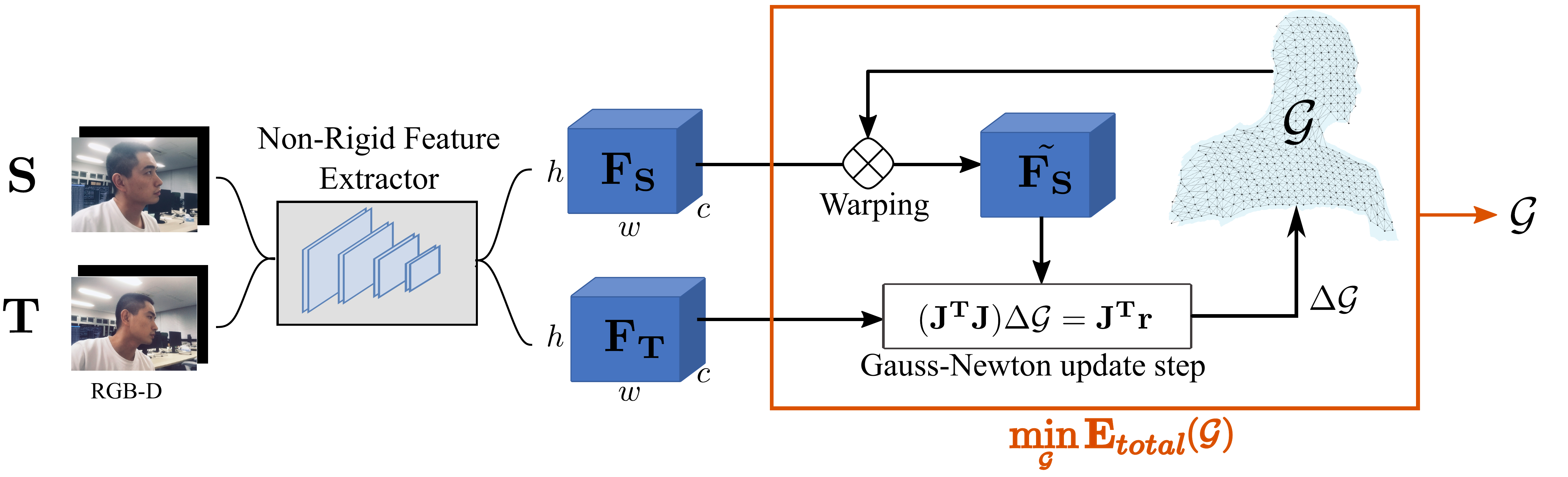}      
    \caption{High-level overview of our non-rigid feature extractor training method.
        Jacobian $\mathbf{J}$'s entries for the feature term (\ref{eqn:feature_term}) can be precomputed according to the inverse composition algorithm. Other entries in Jacobian $\mathbf{J}$ and all entries in residues $\mathbf{r}$ are recomputed in each Gauss-Newton iteration. 
        For simplicity, the geometric fitting term (\ref{eqn:geo_term}) and regularization term (\ref{eqn:prior_term}) are omitted from this figure.}
    \label{fig:feature_learning_pip}
\end{figure*}

\subsection{Total Energy}
To resolve the ambiguity in $Z$ axis, we adopt a projective depth, which is a rough approximation of the point-to-plane constraint, as our \textbf{geometric fitting term}. This term measures the difference between warped depth map $\mathbf{\tilde{D}_S}$ and the depth map of target frame. It is defined as 
\begin{equation}
\label{eqn:geo_term}
\mathbf{E}_{geo}(\mathcal{G}) = \lambda_{g} \sum^{w \times h}_{i=0}  \mathcal{V}_i \cdot||  \mathbf{
    \tilde{D}_S}(i)- \mathbf{D_T}(i)  ||^2
\end{equation}  
Finally, we regularize the shape deformation by the ARAP \textbf{regularization term}, which encourages locally rigid motions. It is defined as
\begin{equation}
\label{eqn:prior_term}
\mathbf{E}_{reg}(\mathcal{G}) =\lambda_{r} \sum^{w \times h}_{i=0} \sum_{j \in \mathcal{N}_i}  \mathcal{E}_{i,j} \cdot|| 
(\mathbf{t}_i -\mathbf{t}_j ) - \mathbf{R}_i 
(\mathbf{t}^{'}_i - \mathbf{t}^{'}_j)   
||^2
\end{equation}Where $\mathcal{N}_i$ denotes node-$i$'s neighboring nodes, and $\mathbf{t}_j^{'},\;\mathbf{t}_j^{'}$ are the positions of $i,\; j$ after the transformation. To summarize the above, we obtain the following energy for non-rigid tracking:
\begin{equation}
\label{eqn:total_energy_ours}
\mathbf{E}_{total}(\mathcal{G}) = \mathbf{E}_{fea}(\mathcal{G}) + \mathbf{E}_{geo}(\mathcal{G}) + \mathbf{E}_{reg}(\mathcal{G})
\end{equation}
The three terms are balanced by $ [ \lambda_f , \lambda_g, \lambda_r ]$. The total energy is then optimized by the Gauss-Newton update steps:
\begin{equation}
\label{eqn:GM_step}
( \mathbf{J^TJ} ) \Delta \mathcal{G} =  \mathbf{J^Tr}
\end{equation}
where $\mathbf{r}$ is the error residue, and $\mathbf{J}$ is the Jacobian of the residue with respect to $\mathcal{G}$. This equation is further solved by the iterative PCG solver.

\subsection{Back-Propagation Through the Two Solvers} \label{section:bp_gm_pcg}
The learning pipeline is shown in Fig. \ref{fig:feature_learning_pip}. We integrate all energy optimization steps into an end-to-end training pipeline. To this end, we need to make both Gauss-Newton and PCG differentiable.
In the Gauss-Newton case, the update steps stop when a specified threshold is reached. Such if-else based termination criteria prevents error back-propagation. 
We apply the same solution as in \cite{deepicn,banet,regnet}, \ie we fix the number of Gauss-Newton iterations. 
In this project, we set this number to a small digit. There are two reasons behind this: 1) For the recursive nature of the Gauss-Newton layer, large iterations number will induce instability to the network training, 2) By limiting the available step, the feature extractor is pushed to produce the features that allow Gauss-Newton solver to make bigger jumps toward the solution. Thus we can achieve faster convergence and robust solving.

Back-propagation through PCG can is done in a different fashion as described in~\cite{bilateral_solver}. 
Equation (\ref{eqn:GM_step}) need to be solved in every Gauss-Newton iteration.  Let's represent $\mathbf{J^TJ}$ by $\mathbf{A}$, $\Delta \mathcal{G}$ by $\mathbf{x}$, and $\mathbf{J^Tr}$ by $\mathbf{b}$, then we get the following iconic equation: 
\begin{equation}
\mathbf{A}  \mathbf{x} =  \mathbf{b}
\end{equation}
Suppose that we have already got the gradient of loss $\mathbb{L}$ \textit{w.r.t} to the solution $\mathbf{x}$ as $\partial{\mathbb{L}}/\partial{\mathbf{x}}$. We want to back propagate that quantity onto $\mathbf{A}$ and $\mathbf{b}$:
\begin{equation}
\label{eqn:gradient_solve}
\frac{\partial{\mathbb{L}}}{\partial{\mathbf{b}}} = \mathbf{A}^{-1}
\frac{\partial{\mathbb{L}}}{\partial{\mathbf{x}}}
\end{equation}
\begin{equation}
\frac{\partial{\mathbb{L}}}{\partial{\mathbf{A}}} = ( -\mathbf{A}^{-1} \frac{\partial{\mathbb{L}}}{\partial{\mathbf{x}}} ) 
(\mathbf{A}^{-1}\mathbf{b})^{\mathrm{T}} = -\frac{\partial{\mathbb{L}}}{\partial{\mathbf{b}}} \mathbf{x}^{\mathrm{T}}
\end{equation}
which means that back-propagating through the linear system only need another PCG solve for Equation (\ref{eqn:gradient_solve}).

\begin{figure}[!t]
    \centering
    \includegraphics[width=1\linewidth]{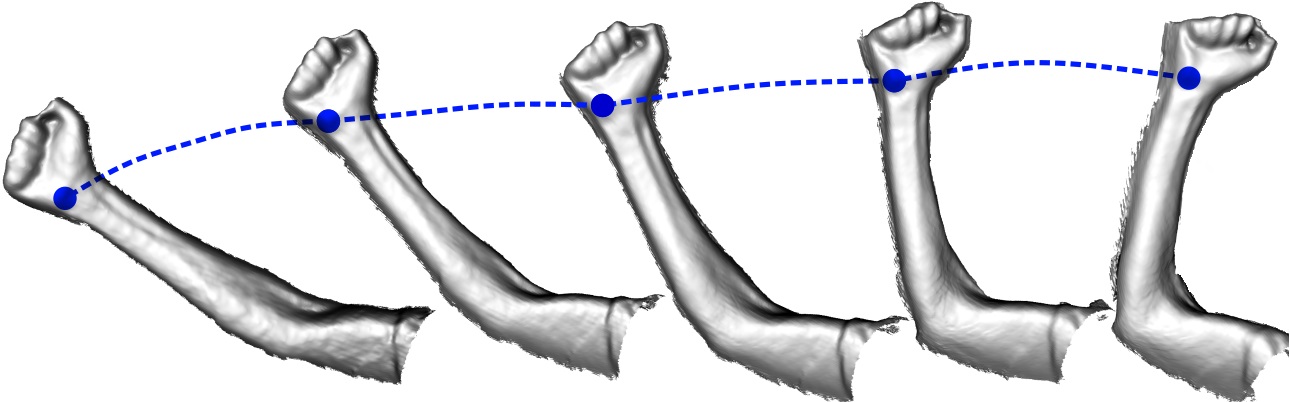}
    \caption{Using our non-rigid tracking and reconstruction method to obtain point-point correspondence. This method can generate accurate correspondence when the motion is small.
    The long term correspondence between distant frames can be obtained by accumulating small inter-frame motions through time and space.}
    \label{fig:correspondence_acquisition}
\end{figure}

\begin{figure*}[t]
    \centering
    \includegraphics[width=1\linewidth]{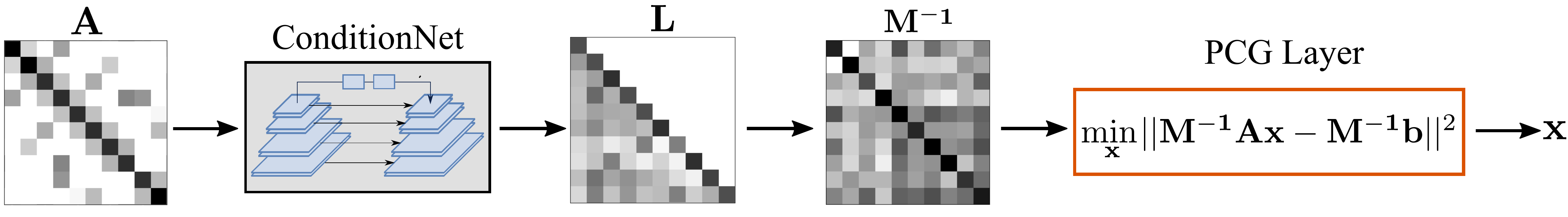}
    \caption{Overview of ConditionNet-Dense. The output $\mathbf{L}$ is the lower triangle matrix of the preconditioner. After a few iterations in the PCG layer, the solution $\mathbf{x}$ is then penalized by the L1 loss (\ref{eqn:loss_pcg}). The whole pipeline can be trained end-to-end. }
    \label{fig:ConditionNet}
\end{figure*}
\subsection{Training Objective \& Data Acquisition}

The method outputs the final deformation graph after a few Gauss-Newton iterations. We apply the L1 flow loss on all the translation vectors $\mathbf{t}_i \in \mathcal{G}$ in the deformation graph 
\begin{equation}
\label{eqn:epe_loss}
\mathbb{L}_{flow} = \sum_{\mathbf{t}_i \in \mathcal{G}} |\mathbf{t}_i - \mathbf{t}_{i,gt}|
\end{equation}
where $\mathbf{t}_{i,gt} \in \mathbb{R}^{3}$ is node-$i$'s ground truth 3D translation vector, \ie the scene flow. 

Collecting $\mathbf{t}_{i,gt}$ is a non-trivial task.  Inspired by Zeng \etal \cite{3dmatch} and Schmidt~\etal~\cite{schmidt2016self}, we realize that the 3D correspondence ground truth can be achieved by running the state-of-the-art tracking and reconstruction methods such as BundleFusion~\cite{dai2017bundlefusion}, for rigid scenes, or DynamicFusion \cite{dynamicfusion}/ VolumeDeform~\cite{volumedeform}, for non-rigidly deforming scenes.
For the rigid training set, we turn to the ScanNet, which contains a large number of indoor sequences with BundleFusion based camera trajectory.  For the non-rigid training dataset, as shown in Fig. \ref{fig:correspondence_acquisition}, we run our geometry based non-rigid reconstruction method (which is similar to DynamicFusion~\cite{dynamicfusion}) on the collected non-rigid sequences. 
We argue that non-rigid feature learning could benefit from rigid scenes. Since the rigid scenes can be considered as a subset of the non-rigid ones, the domain gap is not that huge when we approximate the rigid object surface from a deformable perspective.
Eventually, the feature learning pipeline is pre-trained on ScanNet and fine-tuned on our non-rigid dataset.

\section{Data-Driven Preconditioner} \label{section:conditioner}

Preconditioner $\mathbf{M^{-1}}$ modifies the system $\mathbf{A}  \mathbf{x} =  \mathbf{b}$ to 
\begin{equation}
\label{eqn:conditioned_Ax_b}
\mathbf{M^{-1}A}  \mathbf{x} =  \mathbf{M^{-1}b}
\end{equation} 
which is easier to solve. From the iterative optimization perspective, solving (\ref{eqn:conditioned_Ax_b}) is equal to finding the $\mathbf{x}$ that minimizes the quadratic form
\begin{equation} 
\label{eqn:conditioned_quadratic_form}
\min_{\mathbf{x}} || \mathbf{M^{-1}A}  \mathbf{x} -  \mathbf{M^{-1}b} ||^2
\end{equation}Here, we propose the ConditionNet $\mathcal{C}(\cdot)$ based on neural networks with an encoder and decoder structure to do the mapping:
$$\mathcal{C}(\cdot): \mathbb{R}^{n\times n} \rightarrow  \mathbb{R}^{n\times n}:  \mathbf{A} \rightarrow \mathbf{M^{-1}}$$
A good preconditioner should be a symmetric positive definite (SPD) matrix, otherwise, the PCG can not guarantee to converge. To this end, the ConditionNet first generates the lower triangle matrix $\mathbf{L}$. Then the preconditioner $\mathbf{M^{-1}}$ is computed as
\begin{equation} 
\mathbf{M^{-1} = LL^T}
\label{eqn:LLT}
\end{equation}
Empirically, we apply a hard positive threshold on $\mathbf{M^{-1}}$'s diagonal entries to combat the situations that there exist zero singular values. By doing this, $\mathbf{M^{-1}}$ is ensured to be an SPD matrix in our case.

The matrix density, \ie the ratio of non-zero entries, play an important role in preconditioning. 
On one end, a denser preconditioner has a higher potential to approximate $\mathbf{A^{-1}}$, which is the perfect conditioner, but the matrix inverse itself is time-consuming. On the other end, a sparser matrix is cheaper to achieve while leading to a poor preconditioning effect. 
To examine the trade-off between efficiency and effectiveness, we propose the following three ConditionNet variants. 
They use the same network structure but generate preconditioners with different density, from dense to sparse.


\textbf{ConditionNet-Dense}. As shown in Fig. \ref{fig:ConditionNet}, this one uses full matrix $\mathbf{A}$ as input and generate the dense preconditioner, in which all entries can be non-zero. Intuitively, this model is trying to approximate the perfect conditioner $\mathbf{A^{-1}}$.

\textbf{ConditionNet-Sparse}. This one inputs full matrix $\mathbf{A}$. For the output, a binary mask is applied such that any entry in $\mathbf{L}$ is set to zero if the corresponding entry in $\mathbf{A}$ is also zero. 

\textbf{ConditionNet-Diagonal}. The input and output are the block diagonals of the matrices. 
There are $w\times h$ diagonal blocks and each block is $6\times 6$. Since each block is directly related to a feature in the 2D mesh grid, we reshape the input block diagonal entries to a $[w,h,36]$ volume to leverage such 2D spatial correlations. The output volume is $[w,h,21]$ for the lower triangle matrix $\mathbf{L}$. This model generates the sparsest preconditioner.

\subsection{Self-Supervised Training}

\begin{table*}[!ht]
	\small
	\centering
	\setbox0\hbox{\tabular{@{}l}\small ARAP \endtabular}
	\setbox1\hbox{\tabular{@{}l}\small P2Plane\endtabular}
	\setbox2\hbox{\tabular{@{}l}\small P2Point\endtabular}
	\setbox3\hbox{\tabular{@{}l}\small Color\endtabular}
	\setbox5\hbox{\tabular{@{}l}\small Feature\endtabular}
	\renewcommand{\arraystretch}{1.2} 
	\setlength{\tabcolsep}{3pt} 

	\begin{tabular}{|l||c|c|c|c|c||c|c|c|c||c|c|c|c|}
		\hline
		\multirow{3}{*}{Method} &\multicolumn{5}{c||}{Energy Terms}  &  \multicolumn{4}{c||}{ScanNet (SN)}  &  \multicolumn{4}{c|}{Non-Rigid Dataset (NR)}   \\ \cline {2-14}

		 &\rule{0pt}{\dimexpr\wd0-\normalbaselineskip}  & \rule{0pt}{\dimexpr\wd1-\normalbaselineskip} & \rule{0pt}{\dimexpr\wd2-\normalbaselineskip} & \rule{0pt}{\dimexpr\wd3-\normalbaselineskip}  & \rule{0pt}{\dimexpr\wd5-\normalbaselineskip}  & 
		\multicolumn{4}{c||}{ \makecell{3D EPE (cm) on $\downarrow$ validation/test  \\ on   frame jump }} & 
		\multicolumn{4}{c|}{\makecell{3D EPE (cm) on $\downarrow$ validation/test  \\ on  frame jump } } 
		\\ \cline {7-14}

		& \rotatebox{90}{\rlap{\usebox0}}   & \rotatebox{90}{\rlap{\usebox1}} & \rotatebox{90}{\rlap{\usebox2}} &\rotatebox{90}{\rlap{\usebox3}}   &  \rotatebox{90}{\rlap{\usebox5}}                 &  0$\rightarrow$2    &  0$\rightarrow$4   &  0$\rightarrow$8  & 0$\rightarrow$16   &  0$\rightarrow$2    &  0$\rightarrow$4   &  0$\rightarrow$8  & 0$\rightarrow$16     \\ \hline

        N-ICP-0 &   \checkmark  && & \checkmark &  & 
		3.21/3.65 &  5.03/5.68     &  8.17/9.66  & 15.35/18.65 &  
		2.29/2.3 &   4.08/3.3 & 7.71/6.3& 13.63/12.72    \\ \hline

		N-ICP-1 \cite{dynamicfusion}&  \checkmark & \checkmark&    \checkmark & &  & 
		2.43/2.75 &  4.50/5.38    &  8.11/9.09  & 14.62/17.10   &  
		\textbf{1.49}/1.53 & 2.98/2.71    &   6.61/6.52   &  11.14/12.07      \\ \hline

		N-ICP-2  \cite{zollhofer2014real}&  \checkmark & \checkmark&\checkmark& \checkmark &      &     \textbf{2.04}/2.71&  3.58/4.47   &  6.07/7.89   &  10.36/14.96  &
		1.68/1.70&  3.41/2.60  &   5.50/5.20  &  11.80/10.59     \\ \hline

		Ours (SN)& \checkmark & \checkmark &  &     & \checkmark  & 
		2.10/\textbf{2.60}& \textbf{3.55}/\textbf{4.39}  & \textbf{5.28}/\textbf{6.98} &  \textbf{7.34}/\textbf{10.59}    &  
		1.73/1.60 & 2.77/2.63 & 4.99/5.08 & 7.09/8.32   \\ \hline

		Ours (SN+NR)&\checkmark  & \checkmark& &     &  \checkmark & 
		-- & --   & --    & --    &
		1.55/\textbf{1.34} &  \textbf{2.25}/\textbf{2.23} & \textbf{4.16}/\textbf{4.50} & \textbf{6.47}/\textbf{7.59}      \\ \hline			
	\end{tabular}
	
	\label{tab:tracking_numbers}
	\vspace{0.1cm}
	\caption{3D End point Error (EPE) on ScanNet and our Non-Rigid dataset.  
	The frame jumps shows the index of the indices of the source and target frame. 
	The number of  unkonws in the deformation graph is 1152 ($16\times12\times6$). Ours (SN): trained on ScanNet~\cite{dai2017scannet}. Ours (SN + NR): pretrained on ScanNet and fine-tuned on the Non-Rigid dataset~\cite{deepdeform}.
	}
\end{table*}

The straight forward way to train the ConditionNet is to minimize the condition number $ \kappa (\mathbf{M^{-1}A})={ \lambda _{\max }  /   \lambda _{\min }  } $ \ie the ratio of the maximum and minimum singular value in $\mathbf{M^{-1}A}$. However, the time consuming singular value decomposition (SVD) makes large scale training impractical. 

Instead, we propose the PCG-Loss for training. As shown in Fig.~\ref{fig:ConditionNet} the learned preconditioner $\mathbf{M^{-1}}$ is fed to a PCG layer to minimize (\ref{eqn:conditioned_quadratic_form}) and output the solution $\mathbf{x}$.
Training data generation for the ConditionNet is fully automatic; i.e., no annotations are needed to find the ground truth solution $\mathbf{x}_{gt}$ to equation $\mathbf{A}  \mathbf{x} =  \mathbf{b}$, which we do by running a standard PCG solver.
To obtain $\mathbf{x}_{gt}$, the standard PCG is executed as many iterations as possible till convergence. Then the L1 PCG-Loss is applied on the predicted solution
\begin{equation}
\mathbb{L}_{pcg}= |\mathbf{x} - \mathbf{x}_{gt}| 
\label{eqn:loss_pcg}
\end{equation}
The training samples, \ie the [$\mathbf{A}$, $\mathbf{b}$] pairs, are collected from the Gauss-Newton update step in Eqn. (\ref{eqn:GM_step}). 

During the training phase, we limit the number of available iterations in the PCG layer. This is to encourage the ConditionNet to generate a better preconditioner that achieves the same solution while using fewer steps. At the early phase of the training, the PCG layer with limited iterations does not guarantee a good convergence. The back-propagation strategy described in Section \ref{section:bp_gm_pcg} can not be applied here, because incomplete solving results in wrong gradient. Instead, we directly flow the gradient through all PCG iterations for ConditionNet training.

We train ConditionNet and the non-rigid feature extractor separately. They are used together at the testing phase.

\begin{figure*}[!ht]

	\setbox0\hbox{\tabular{@{}c} \small Scene  \endtabular}
	\setbox1\hbox{\tabular{@{}c} \textbf{\textit{ \hspace{1.2cm} bear} } \endtabular}
	\setbox2\hbox{\tabular{@{}c} \textbf{\textit{ \hspace{1cm} pillow} } \endtabular}
	\setbox3\hbox{\tabular{@{}c} \textbf{\textit{ \hspace{1.4cm} adult} } \endtabular}
	\setbox4\hbox{\tabular{@{}c} \textbf{\textit{ \hspace{0.8cm} nameko} } \endtabular}
	\setbox5\hbox{\tabular{@{}c} \textbf{\textit{ \hspace{1.4cm} clothes} } \endtabular}

	\centering
	\begingroup
	\setlength{\tabcolsep}{2pt} 
	\renewcommand{\arraystretch}{1} 
	\small
	
	\begin{tabular}{|c c|c|cc|cc|cc|cc|cc|}
		\hline 
		\rule{0pt}{\dimexpr\wd0-\normalbaselineskip} & \multicolumn{4}{|c|}{  \makecell{Mesh\red{$^1$} and\\   Masked Color Image }  }  & \multicolumn{2}{c| }{  \multirow{2}{*}{Initial Alignment\red{$^2$}}} &\multicolumn{6}{c|}{ \makecell{  Final Alignment, \\   Transformed Source Mesh, and Alignment Error\red{$^3$}   }  } 	
		\\ \cline {2-5} 		\cline {8-13}
		
&\multicolumn{2}{|c|}{Source} & \multicolumn{2}{c| }{ Target} & \multicolumn{2}{c| }{} & \multicolumn{2}{c|}{N-ICP-1~\cite{dynamicfusion}} & \multicolumn{2}{c| }{N-ICP-2~\cite{zollhofer2014real}} & \multicolumn{2}{c|}{Ours} \\ 
		\hline

		\rule{0pt}{\dimexpr\wd1-\normalbaselineskip}&
		\multicolumn{2}{|c|} {\includegraphics[width=0.09\linewidth]{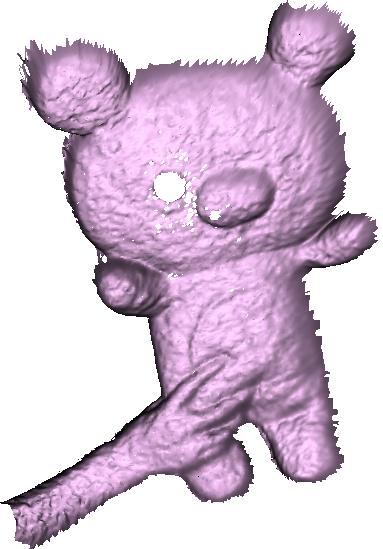} }& 
		\multicolumn{2}{c|} {\includegraphics[width=0.08\linewidth]{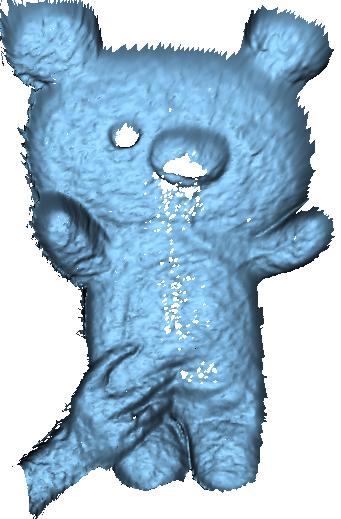}} & 
		\multicolumn{2}{c|} {\includegraphics[width=0.09\linewidth]{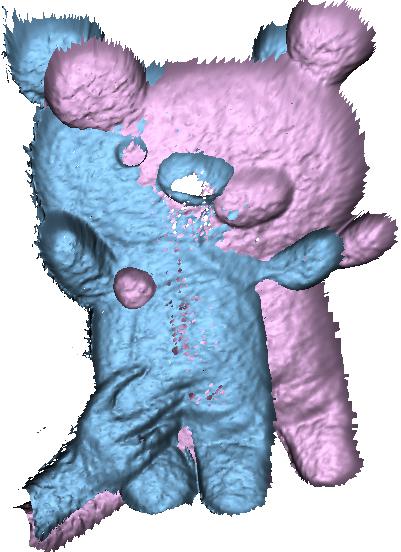}} & 
		\multicolumn{2}{c|} {\includegraphics[width=0.09\linewidth]{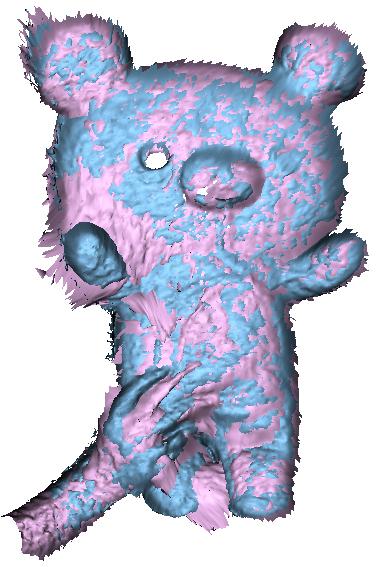}} & 
		\multicolumn{2}{c|} {\includegraphics[width=0.095\linewidth]{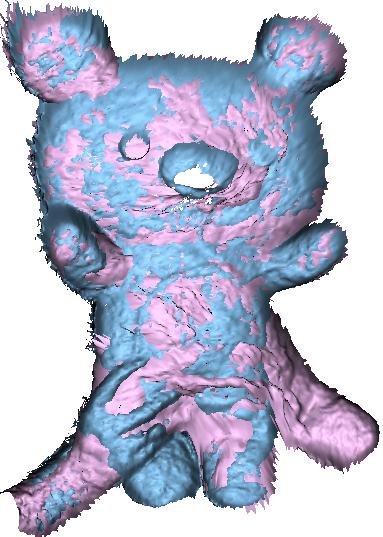}} & 
		\multicolumn{2}{c|} {\includegraphics[width=0.09\linewidth]{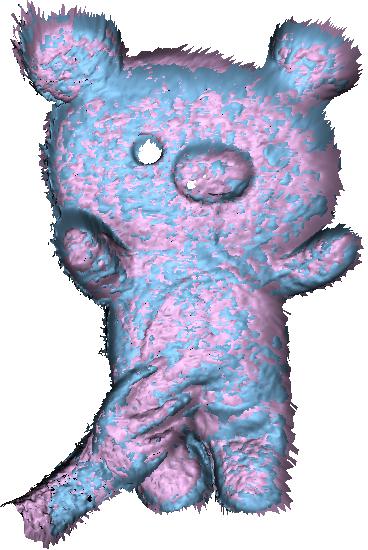}} \\

		\rotatebox{90}{\rlap{\usebox1}}&
		\multicolumn{2}{|c|}{\includegraphics[width=0.065\linewidth]{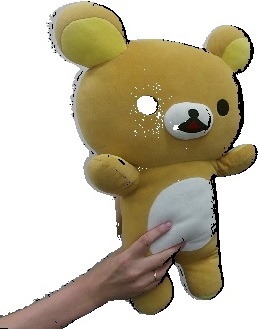}}& 
		\multicolumn{2}{c|}{\includegraphics[width=0.055\linewidth]{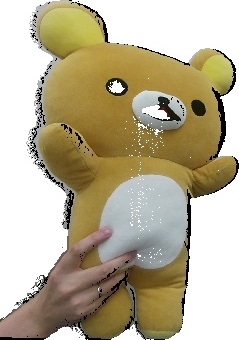}}& 
		\includegraphics[width=0.055\linewidth]{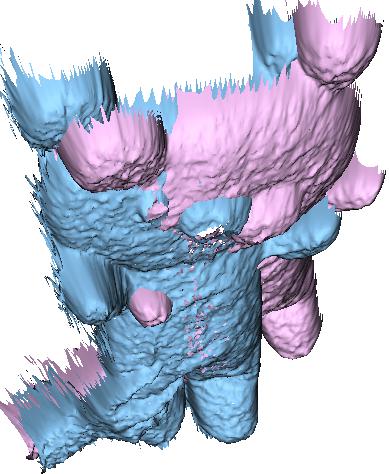}&
		\includegraphics[width=0.055\linewidth]{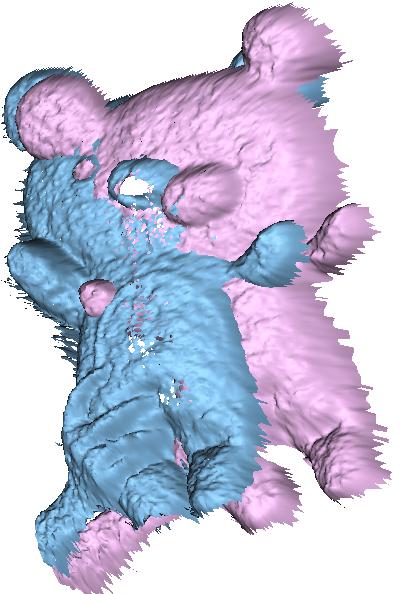}&    
		\includegraphics[width=0.055\linewidth]{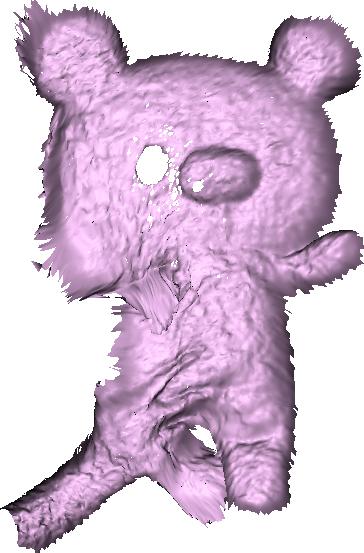}&
		\includegraphics[width=0.055\linewidth]{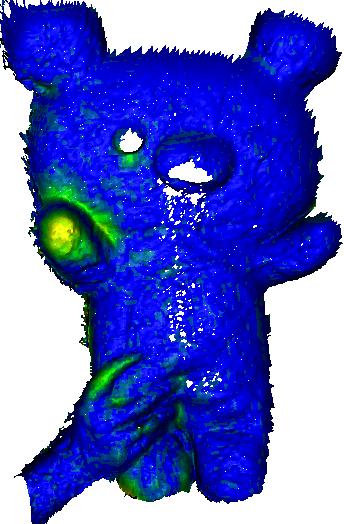} &
		\includegraphics[width=0.06\linewidth]{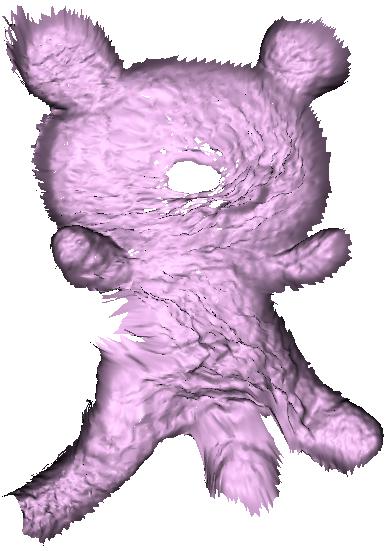}&
		\includegraphics[width=0.055\linewidth]{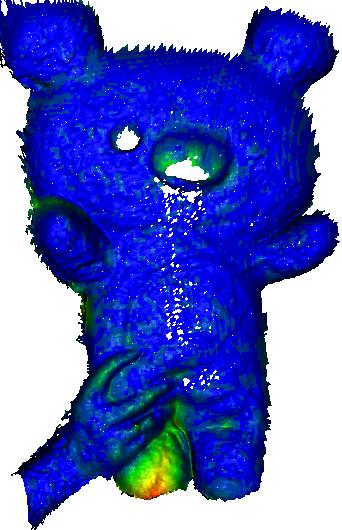}& 
		\includegraphics[width=0.055\linewidth]{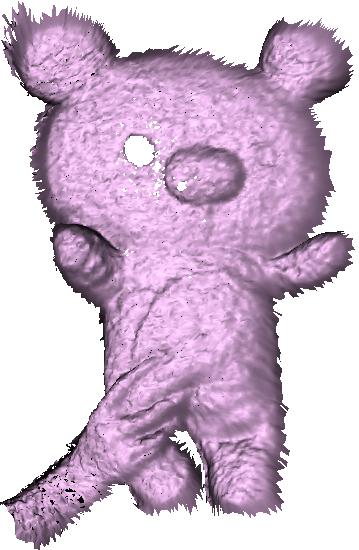}&
		\includegraphics[width=0.055\linewidth]{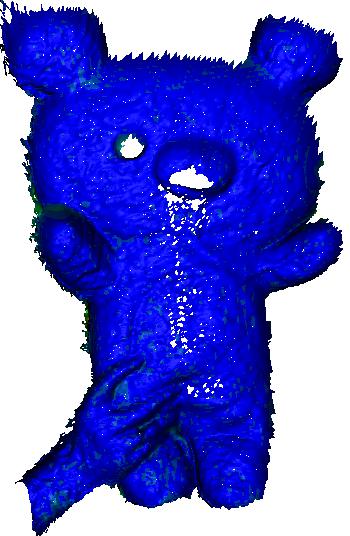}  \\ \hline

		\rule{0pt}{\dimexpr\wd2-\normalbaselineskip}&
		\multicolumn{2}{|c|} {\includegraphics[width=0.11\linewidth]{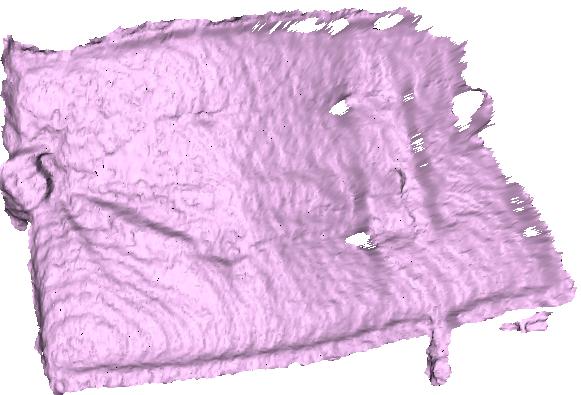} }& 
		\multicolumn{2}{c|} {\includegraphics[width=0.11\linewidth]{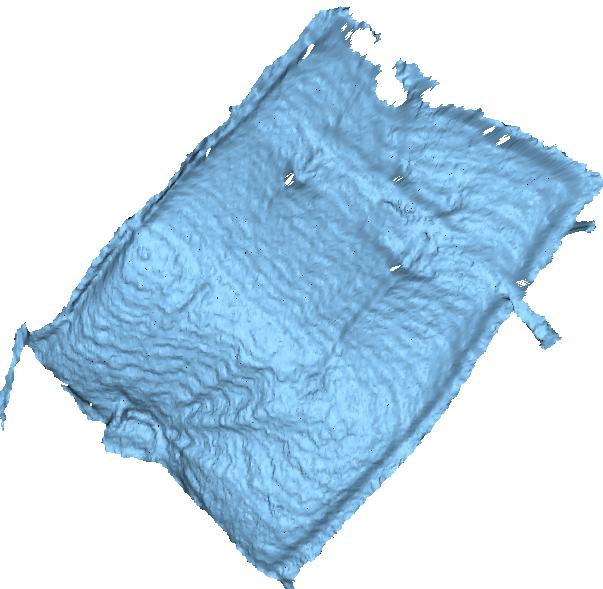}} & 
		\multicolumn{2}{c|} {\includegraphics[width=0.11\linewidth]{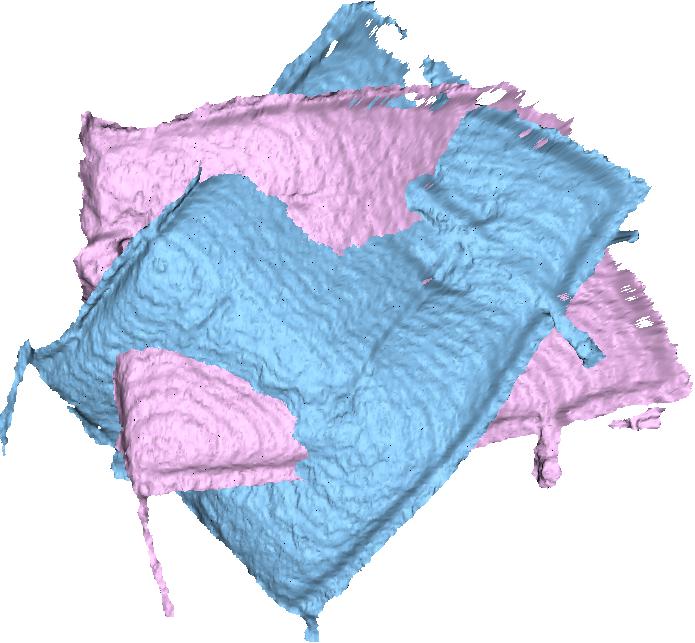}} & 
		\multicolumn{2}{c|} {\includegraphics[width=0.11\linewidth]{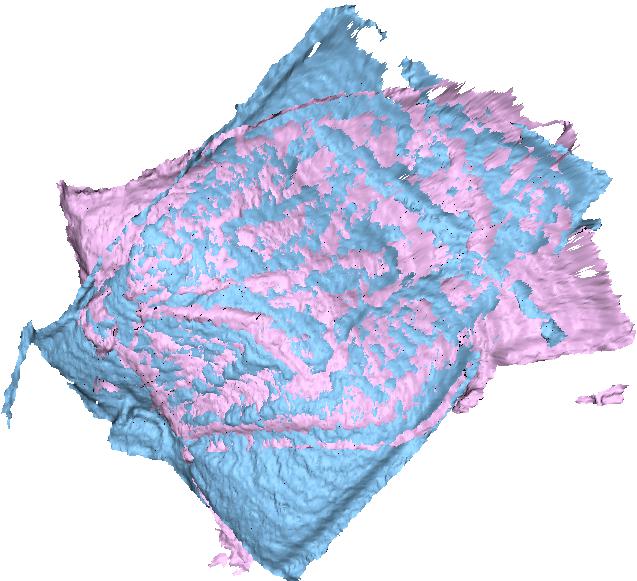}} & 
		\multicolumn{2}{c|} {\includegraphics[width=0.11\linewidth]{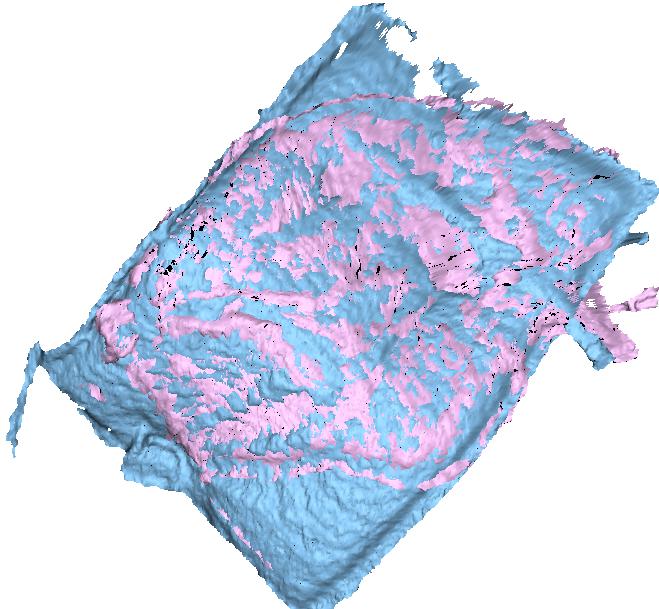}} & 
		\multicolumn{2}{c|} {\includegraphics[width=0.11\linewidth]{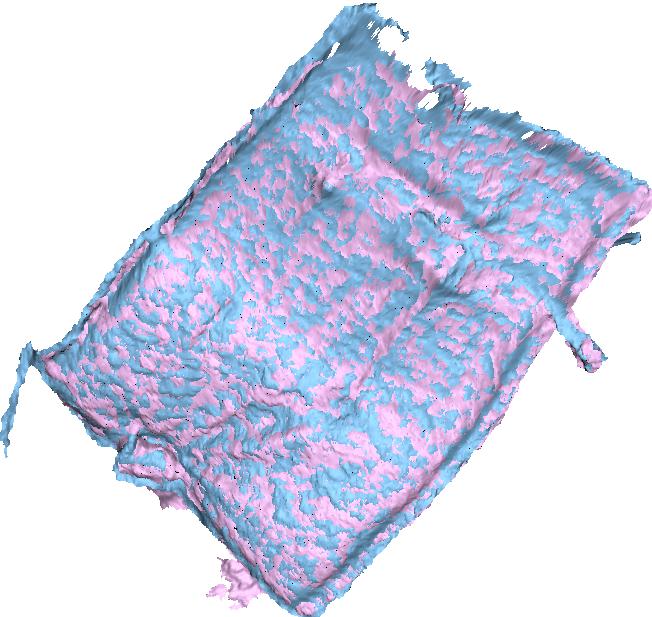}} \\

		\rotatebox{90}{\rlap{\usebox2}}&
		\multicolumn{2}{|c|}{\includegraphics[width=0.065\linewidth]{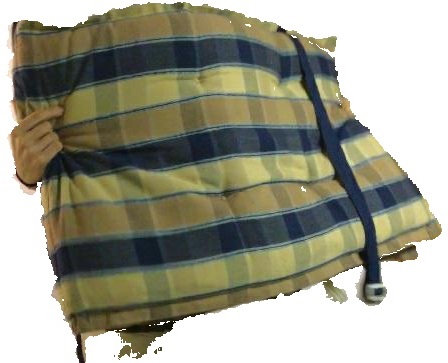}}& 
		\multicolumn{2}{c|}{\includegraphics[width=0.055\linewidth]{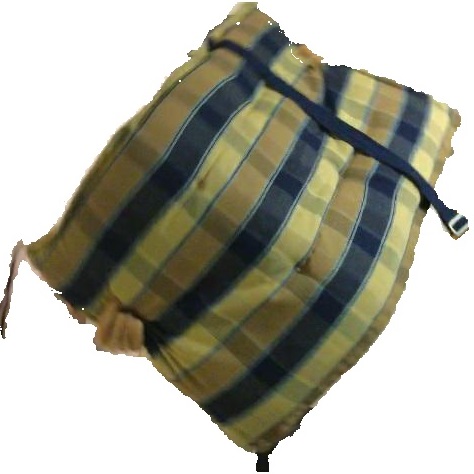}}& 
		\includegraphics[width=0.055\linewidth]{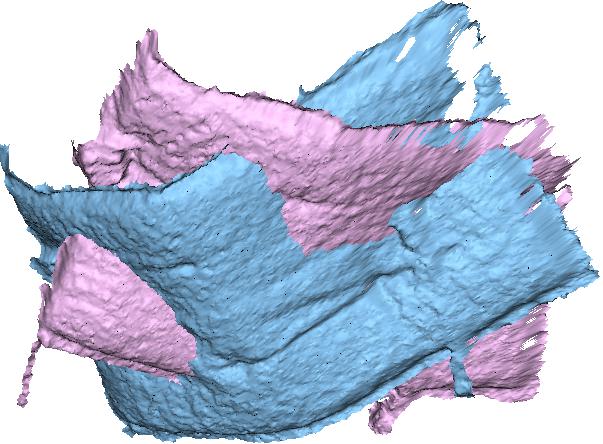}&
		\includegraphics[width=0.055\linewidth]{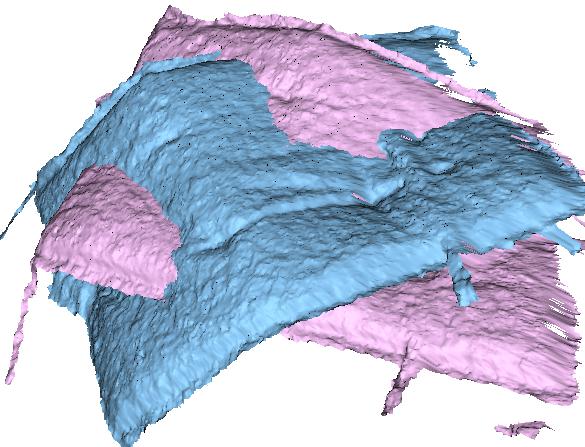}&    
		\includegraphics[width=0.055\linewidth]{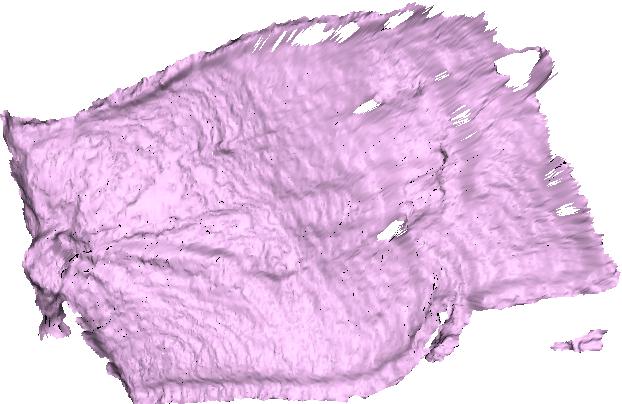}&
		\includegraphics[width=0.055\linewidth]{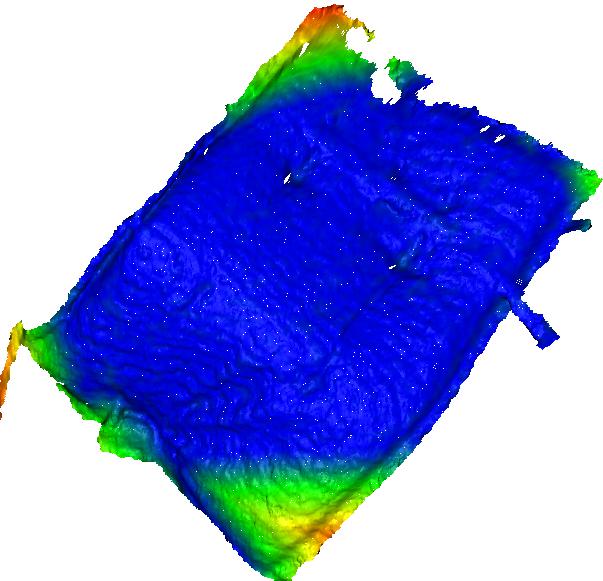} &
		\includegraphics[width=0.06\linewidth]{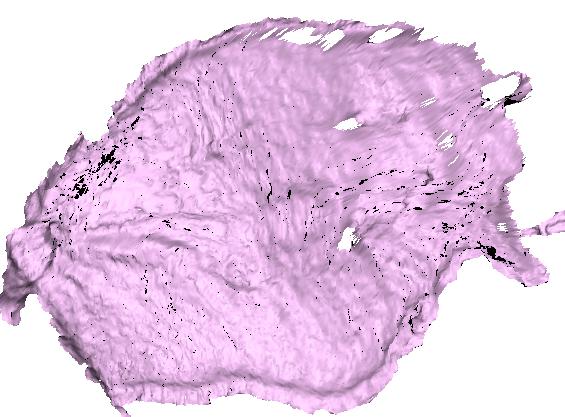}&
		\includegraphics[width=0.055\linewidth]{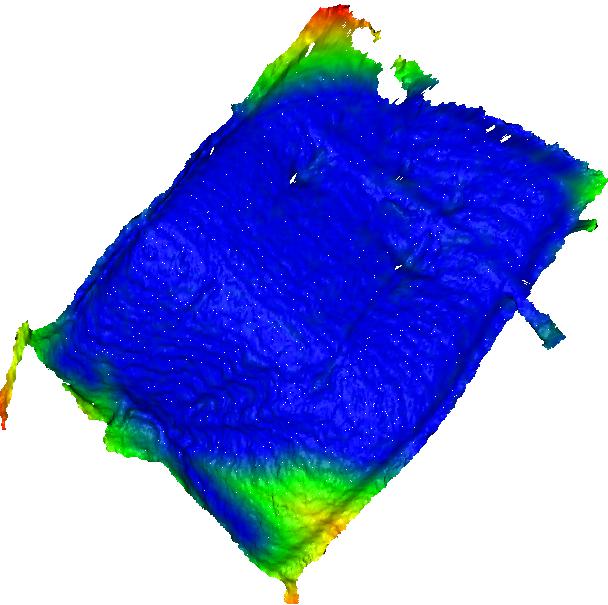}& 
		\includegraphics[width=0.055\linewidth]{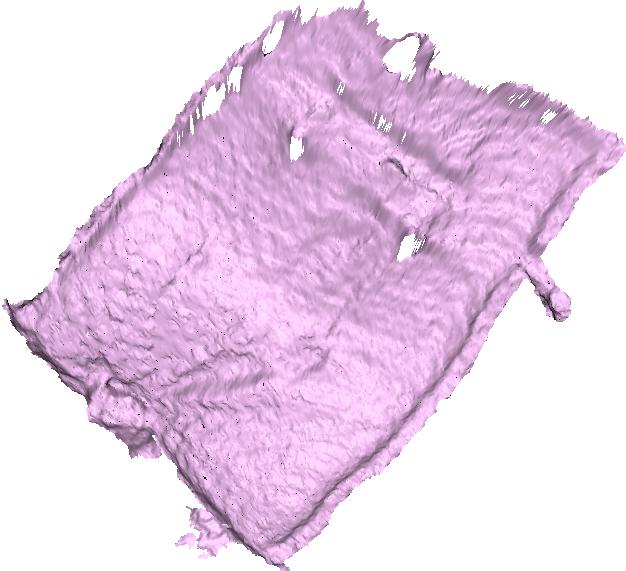}&
		\includegraphics[width=0.055\linewidth]{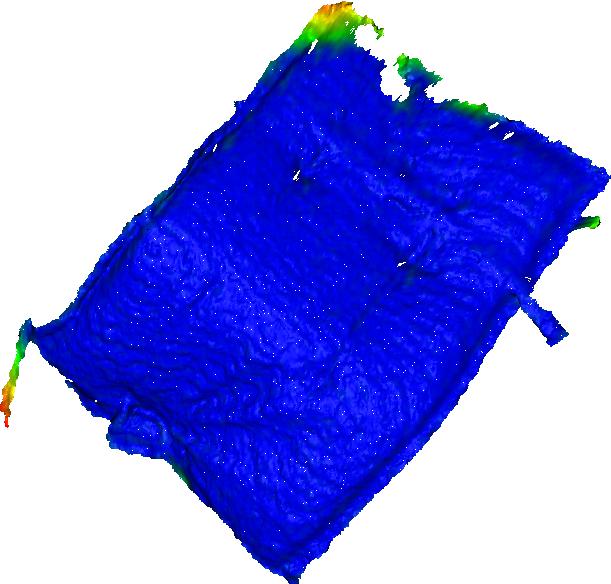}  \\ \hline

		\rule{0pt}{\dimexpr\wd3-\normalbaselineskip}&
		\multicolumn{2}{|c|} {\includegraphics[width=0.12\linewidth]{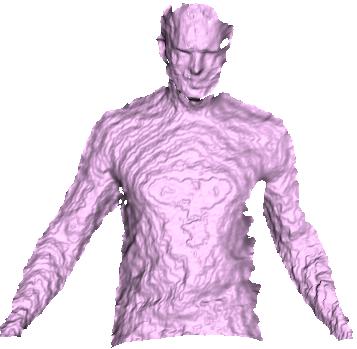} }& 
		\multicolumn{2}{c|} {\includegraphics[width=0.11\linewidth]{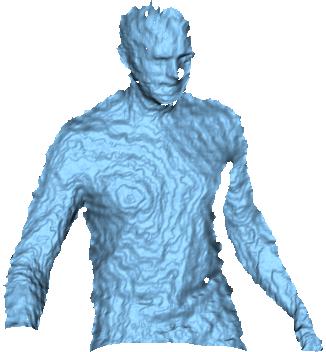}} & 
		\multicolumn{2}{c|} {\includegraphics[width=0.135\linewidth]{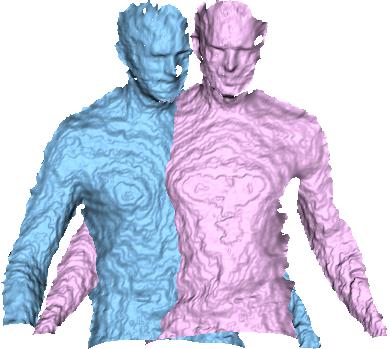}} & 
		\multicolumn{2}{c|} {\includegraphics[width=0.11\linewidth]{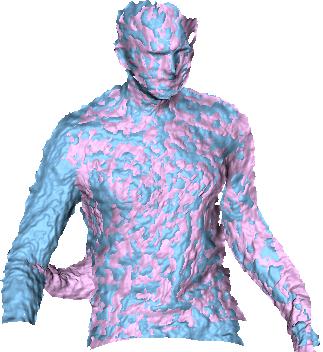}} & 
		\multicolumn{2}{c|} {\includegraphics[width=0.115\linewidth]{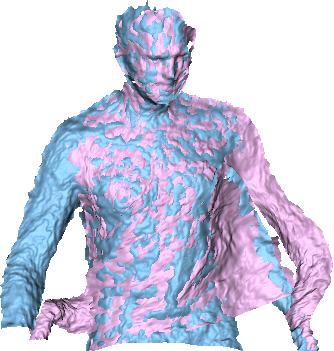}} & 
		\multicolumn{2}{c|} {\includegraphics[width=0.11\linewidth]{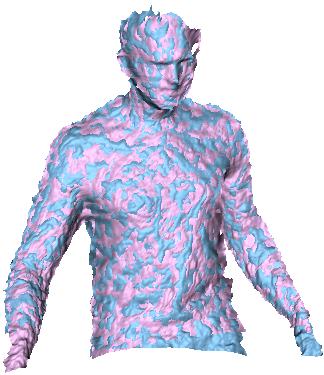}} \\

		\rotatebox{90}{\rlap{\usebox3}}&
		\multicolumn{2}{|c|}{\includegraphics[width=0.075\linewidth]{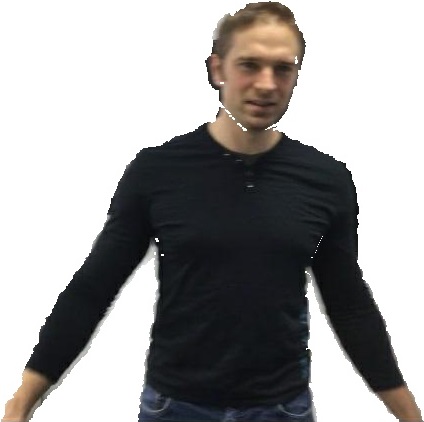}}& 
		\multicolumn{2}{c|}{\includegraphics[width=0.07\linewidth]{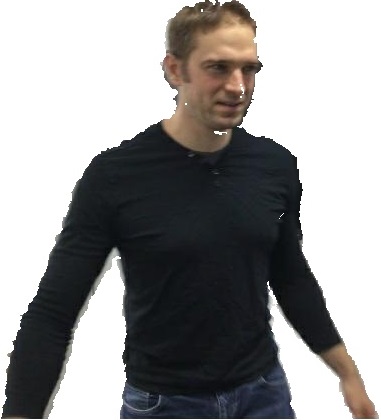}}& 
		\includegraphics[width=0.065\linewidth]{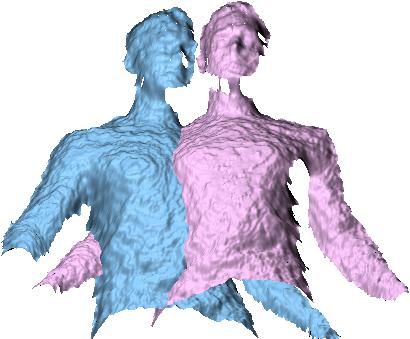}&
		\includegraphics[width=0.065\linewidth]{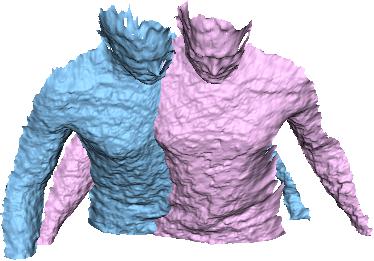}&    
		\includegraphics[width=0.065\linewidth]{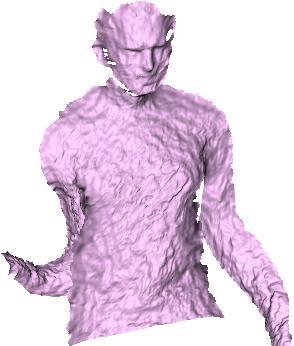}&
		\includegraphics[width=0.065\linewidth]{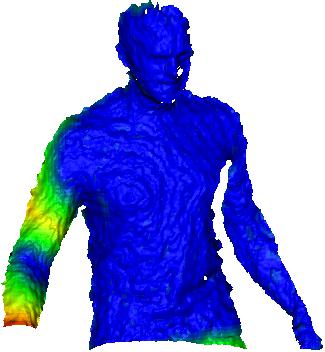} &
		\includegraphics[width=0.065\linewidth]{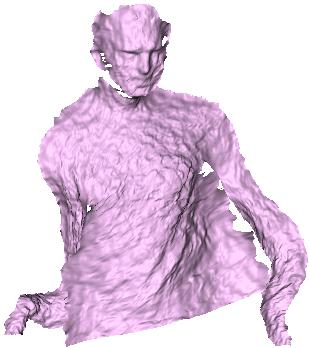}&
		\includegraphics[width=0.065\linewidth]{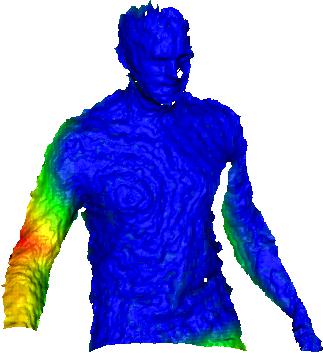}& 
		\includegraphics[width=0.065\linewidth]{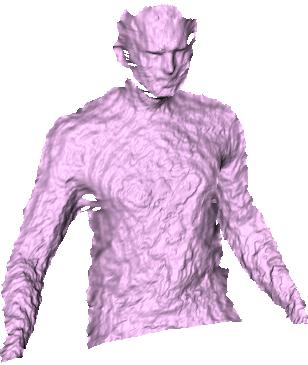}&
		\includegraphics[width=0.065\linewidth]{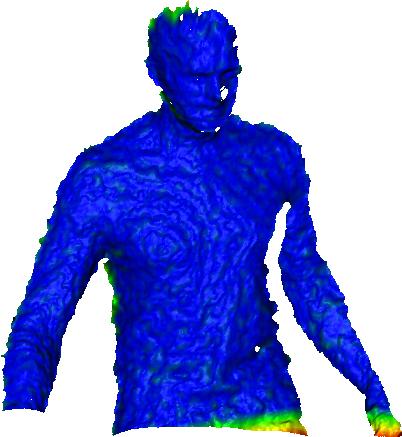}  \\ \hline

		\rule{0pt}{\dimexpr\wd5-\normalbaselineskip}&
		\multicolumn{2}{|c|} {\includegraphics[width=0.135\linewidth]{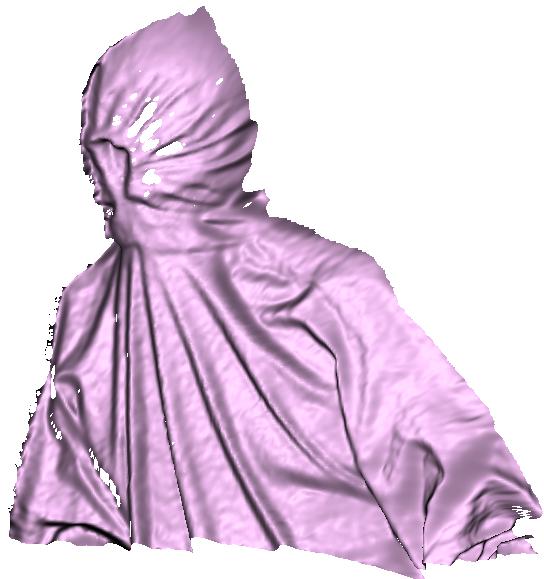} }& 
		\multicolumn{2}{c|} {\includegraphics[width=0.135\linewidth]{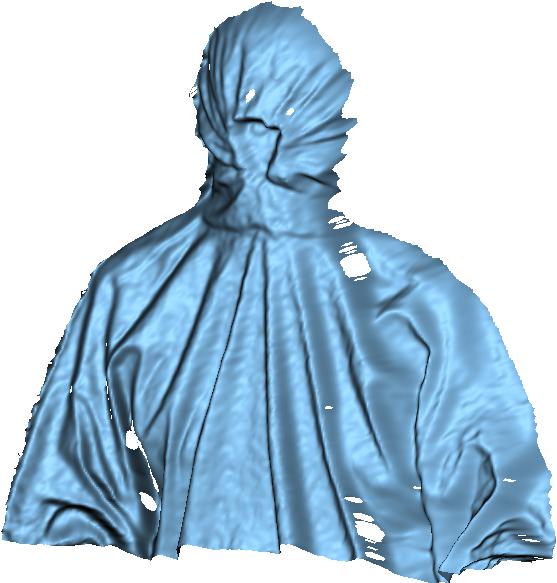}} & 
		\multicolumn{2}{c|} {\includegraphics[width=0.135\linewidth]{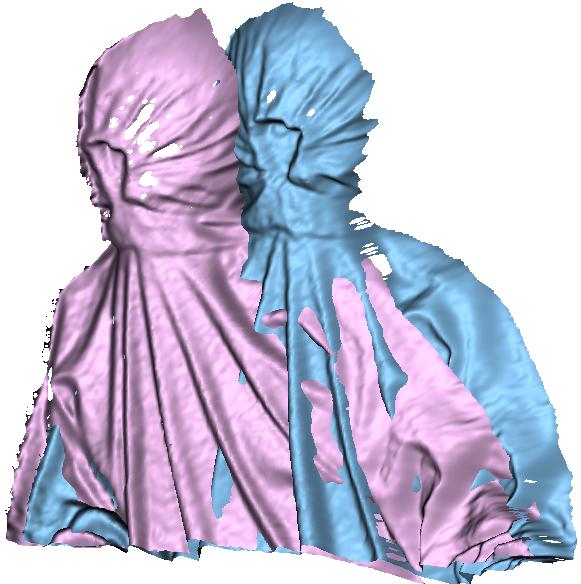}} & 
		\multicolumn{2}{c|} {\includegraphics[width=0.135\linewidth]{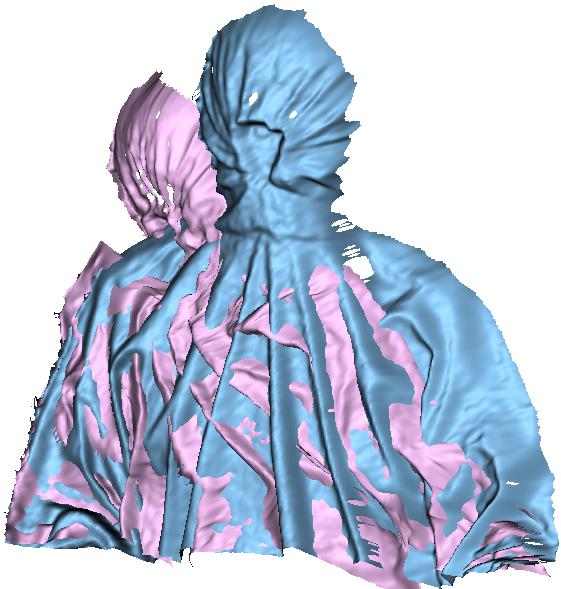}} & 
		\multicolumn{2}{c|} {\includegraphics[width=0.135\linewidth]{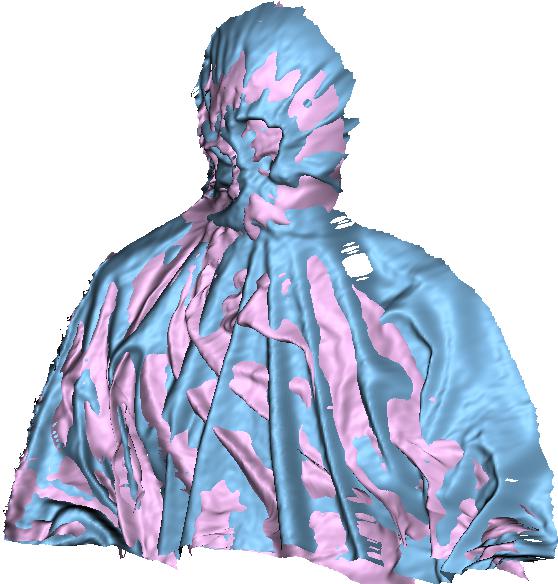}} & 
		\multicolumn{2}{c|} {\includegraphics[width=0.135\linewidth]{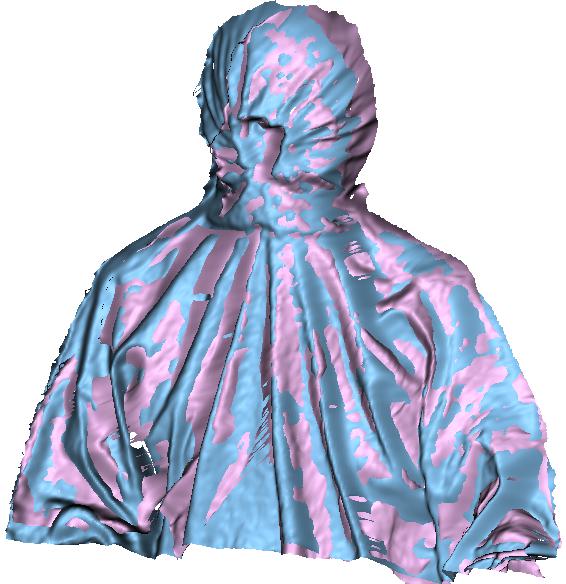}} \\

		\rotatebox{90}{\rlap{\usebox5}}&
		\multicolumn{2}{|c|}{\includegraphics[width=0.082\linewidth]{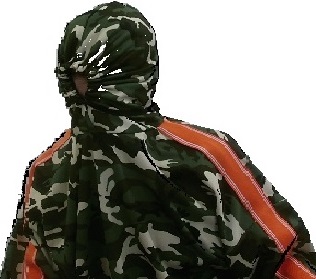}}& 
		\multicolumn{2}{c|}{\includegraphics[width=0.082\linewidth]{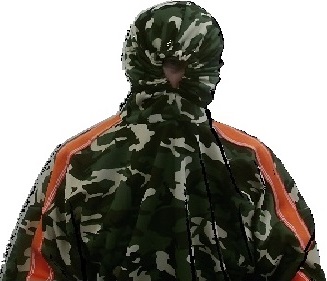}}& 
		\includegraphics[width=0.072\linewidth]{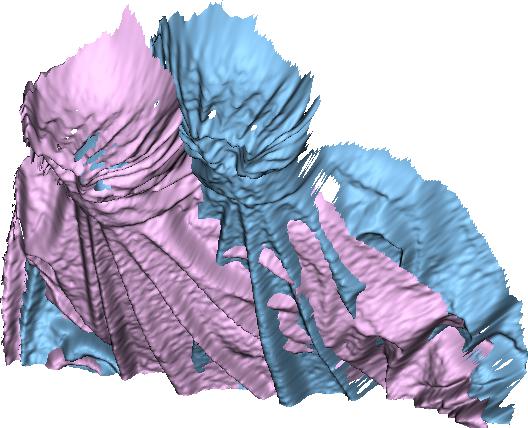}&
		\includegraphics[width=0.072\linewidth]{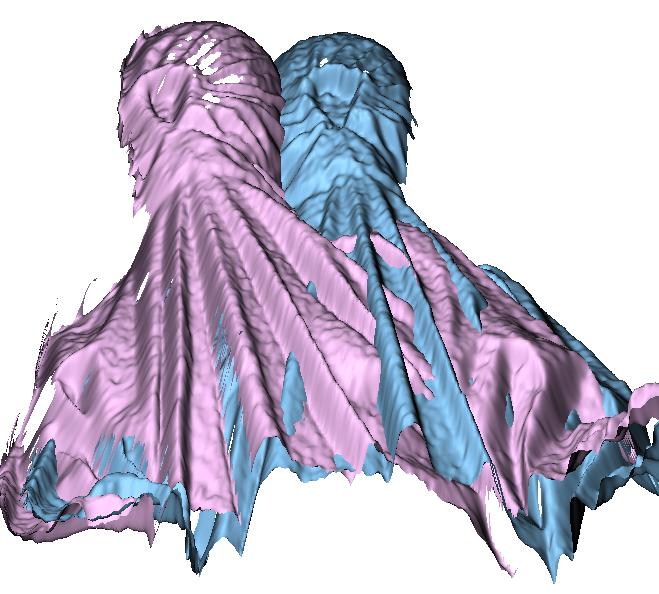}&    \includegraphics[width=0.072\linewidth]{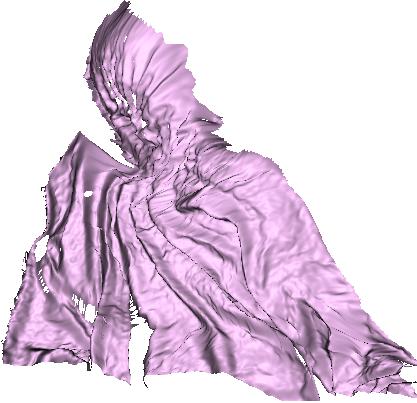}&
		\includegraphics[width=0.072\linewidth]{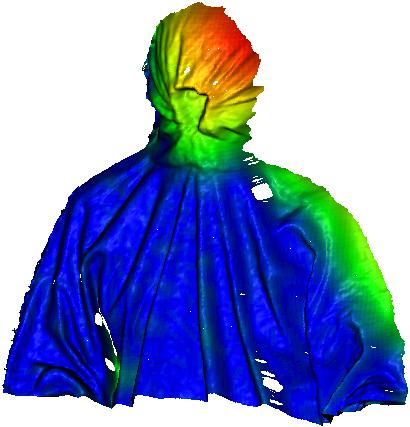} &
		\includegraphics[width=0.072\linewidth]{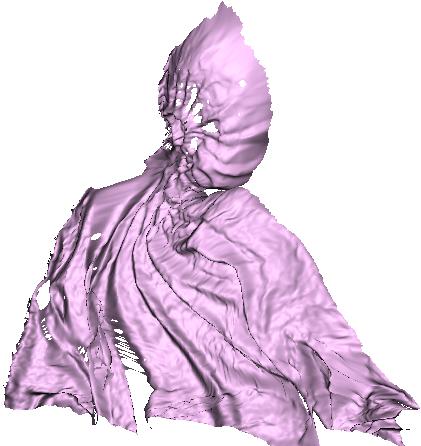}&
		\includegraphics[width=0.072\linewidth]{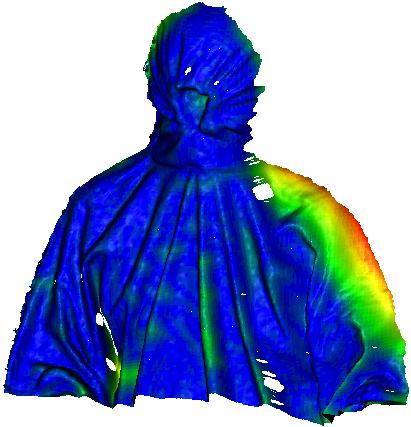}& 
		\includegraphics[width=0.072\linewidth]{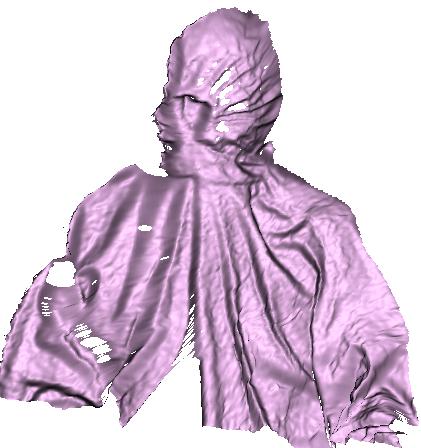}&
		\includegraphics[width=0.072\linewidth]{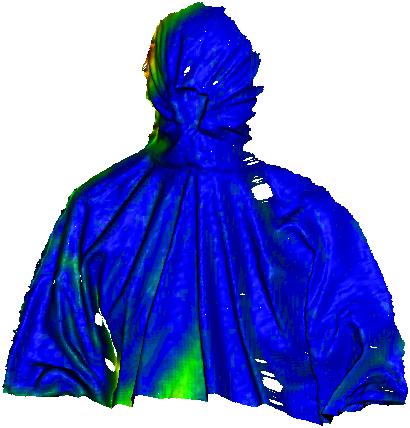}  \\ \hline
		
	\end{tabular}
	\endgroup
	
	\caption {Frame-Frame tracking results.  
	    \red{$^1$} Meshes are constructed from depth images. Depth images are preprocessed by the bilateral filter to reduce observation noise.
        \red{$^2$} Initial alignment is done by simply setting the camera poses of both frames to identity.
        \red{$^3$} The alignment error (hotter means larger) measures the point to point distance between target mesh and the transformed source mesh.
	}
	\label{fig:registration}
\end{figure*}

\section{Experiments}
\paragraph{Implementation Details:}
The resolution of the deformation graph is $16\times12$. 
Empirically, the weighting factor $ [ \lambda_f , \lambda_g, \lambda_r ]$ in the energy function (\ref{eqn:total_energy_ours}) are set to $ [ 1, 0.5, 40]$. The number of Gauss-Newton iterations is 3 for non-rigid feature extractor training.
The number of PCG iterations is 10 for ConditionNet Training.
We implement our networks using the publicly available Pytorch framework and train it with Tesla P100 GPUs. 
We trained all the models from scratch for 30 epochs,with a mini-batch size of 4 using Adam~\cite{adam} optimizer, where  $\beta_1=0.9$,  $\beta_1=0.999$. We used an initial learning rate of 0.0001 and halve it every 1/5 of the total iterations.
\subsection{Datasets}
\paragraph{ScanNet}
ScanNet \cite{dai2017scannet} is a large-scale RGBD video dataset containing 1,513 sequences in 706 different scenes. The sequences are captured by iPad Mounted RGBD sensors that provide calibrated depth-color pairs of VGA resolution.  The 3D camera poses are based on BundleFusion \cite{dai2017bundlefusion}.
The 3D dense motion ground truth on the ScanNet is obtained by projecting point cloud via depth and 6-Dof camera pose. We apply the following filtering process for training data. To narrow the domain gap with the non-rigid dataset, we filter out images if more than 50\% of the pixels have the invalid depth or depth values larger than 2 meters. To avoid image pairs with large pose error, we filter image pairs with a large photo-consistency error. Finally, we remove the image pairs with less than 50\% ``covisibility", \ie the percentage of the pixels that are visible from both images. Similarly, the sequences are subsampled using the intervals [2, 4, 8, 16]. We use 60k frame pairs in total and split train/valid/test as 8/1/1.

\paragraph{Non-Rigid Dataset} 
We use the non-rigid dataset from Alja{\v{z}}~\etal~\cite{deepdeform} which consists of 400 non-rigidly deforming scenes, over 390,000 RGB-D frames. A variety of deformable objects are captured including adults, children, bags, clothes, and animals, etc. The distance of the objects to the camera center lies in the range [0.5m, 2.0m]. Depending on the complexity of the scene, the foreground object masks are either obtained by a simple depth threshold or depth map aided human annotation. We run our tracking and reconstruction method to obtain the ground truth non-rigid motions. We remove the drifted sequences by manually checking the tracking quality of the reconstructed model. The example of this dataset can be found in the paper~\cite{deepdeform}. Similarly to the rigid case, we sub-sample the sequences using the frame jumps [2, 4, 8, 16] to simulate the different magnitude of non-rigid deformation. For data-augmentation, we perform horizontal flips, random gamma, brightness, and color shifts for input frame pairs. Finally, we got 8.5k frame pairs in total and split train/valid/test as 8/1/1.

\subsection{Non-Rigid Tracking Evaluation}

\paragraph{Baselines} We implement a few variants of the non-rigid ICP (N-ICP) methods. They apply different energy terms as shown in Tab. 1.
Among them, N-ICP-1 is our implementation of the method DynamicFusion~\cite{dynamicfusion}, and  N-ICP-2 is our implementation for the method described in~\cite{zollhofer2014real}. The original two papers are focusing on the model to frame tracking problems where the model is either reconstructed on-the-fly or pre-defined. Here all baselines are deployed for the frame-frame tracking problem. Ours first optimizes the feature fitting term based objective (\ref{eqn:total_energy_ours}) to get the coarse motion and then refine the graph with the classic point-to-plane constraints using the raw depth maps.

\paragraph{Quantitative Results} The quantitative results on the ScanNet dataset and the non-rigid dataset can be found in Table 1. The estimated motions are evaluated using the 3D End-Point-Error (EPE) metric. 
On ScanNet, Ours(SN) achieves overall better performance than the other N-ICP baselines, especially when the motions are large (\eg on 0$\rightarrow$8 and 0$\rightarrow$16 frame jump). Note that the ScanNet pre-trained model Ours(SN) even achieves better results than the classic N-ICPs on the non-rigid dataset, indicating a good generalization ability of the learned non-rigid feature, which makes sense considering that the learnable CNN model focuses only on the feature extraction part, and the using of classic optimizer disentangle the direct mapping from images to motion. It also proves the assumption that the rigid and non-rigid surfaces lie in quite close domains. The fine-tuned model Ours(SN+NR) on the non-rigid dataset further improved these numbers. 

\paragraph{Qualitative Results}
Fig. \ref{fig:registration} shows the frame-frame tracking results on the non-rigid frame pairs. We selected the frame pairs with relatively large non-rigid motions. N-ICP-1 and N-ICP-2 have trouble dealing with these motions and converged to bad local minimums. 
Our method manages to converge to the global solutions on these challenging cases.
For instance, the \textbf{\textit{clothes}} scene in Fig.~\ref{fig:registration} is an especially challenging case for classic non-rigid ICP methods because the point-to-plane term has no chance to slide over the zigzag clothes surface which contains multiple folds, and the color consistency term could also be easily confused by the repetitive camouflage textures of the \textbf{\textit{clothes}}. The learned features show an advantage for capturing high order deformation on those cases.

\subsection{Preconditioning Results}

We randomly collected 10K [$\mathbf{A}$, $\mathbf{b}$] pairs from different
iterations the Gauss-Newton step. We split them to
train/valid/test according to the ratio of 8/1/1. We compare
with 3 PCG baselines: w/o preconditioner,
the standard block-diagonal preconditioner, and the Incomplete
Cholesky factorization based preconditioner. We also
show the ablation studies on three of the ConditionNet variants:
Diagonal, Sparse and Dense. Fig. \ref{fig:pcg_convergence} shows the PCG
steps using different preconditioners. The learned preconditioner
outperforms the classic ones by a large margin. Tab. 2
shows PCG’s solving results using different preconditioners.
All learned preconditioners significantly reduced the condition numbers.
ConditionNet-Dense achieves the best convergence
rate and the least overall solving time.

\begin{table}[!t]
\footnotesize
\centering
\renewcommand{\arraystretch}{1.2}  
\setlength{\tabcolsep}{4pt}  
\begin{tabular}{|l|c|c|c|c|}
\hline
Preconditioner      & \textit{density}  & $\kappa$  &   \textit{iters}  & \textit{time (ms)}\\ \hline
None                &--                 & 3442.18   & 46                &  33.43 \\ \hline
Block-Diagonal      & 0.46\%        & 541.52     & 44                & 31.34 \\ \hline
Incomplete Cholesky &1.52            & 379.82    & 37                & 28.42 \\ \hline
ConditionNet-Diagonal (ours) &0.46\%        & 93.55     &21 & 12.38 \\ \hline 
ConditionNet-Sparse (ours)   & 1.52\% & 125.81  & 23 &  17.80 \\ \hline
ConditionNet-Dense (ours)    &  100.\% &\textbf{34.90}   & \textbf{13} &   \textbf{10.32} \\ \hline
\end{tabular}
\label{pcg_numbers}
\vspace{0.1cm}
\caption{PCG solving results using different preconditioners (residue threshold of convergence: $10^{-6}$). \textit{density}: density of preconditioner. $\kappa$: condition number of the modified linear system. \textit{iters}: total steps for convergence. \textit{time(ms)}: time of solving. All numbers are obtained with Pytorch-GPU implementation.
}  
\end{table}

\section{Conclusion and Discussion}
In this work, we present an end-to-end learning approach for non-rigid RGB-D tracking.
Our core contribution is the learnable optimization approach which improves both robustness and convergence by a significant margin. The experimental results show that the learned non-rigid feature significantly improves the convergence of Gauss-Newton solver for the frame-frame non-rigid tracking. In addition, our method increases the PCG solver's convergence rate by predicting a good preconditionier. Overall, the learned preconditioner requires  2 to 3 times fewer iterations until convergence.

While we believe  this results are very promising and can lead to significant practical improvements in non-rigid tracking and reconstruction frameworks, there are several major challenges are yet to be addressed: 
1) The proposed non-rigid feature extractor adopted plain 2D convolution kernels, which are potentially not the best option to handle 3D scene occlusions. One possible research avenue is to directly extract non-rigid features from 3D point clouds or mesh structures using the point-based architectures~\cite{pointnet++}, or even graph convolutions~\cite{geometric_graph_CNN}.
2) Collecting dense scene flow using DynamicFusion for real-world RGB-D video sequence is expensive (\ie~segmentation and outlier removal can become painful processes).
The potential solution is learning on synthetic datasets. (\eg using graphics simulations where the dense motion ground truth is available).

\section*{Acknowledgements}
This work was also supported by a TUM-IAS Rudolf M\"o{\ss}bauer Fellowship, the ERC Starting Grant \textit{Scan2CAD} (804724), and the German Research Foundation (DFG) Grant \textit{Making Machine Learning on Static and Dynamic 3D Data Practical}, as well as the JST CREST Grant Number JPMJCR1403, and partially supported by JSPS KAKENHI Grant Number JP19H01115. YL was supported by the Erasmus+ grant during his stay at TUM.
We thank Christopher Choy, Atsuhiro Noguchi, Shunya Wakasugi, and Kohei Uehara for helpful discussion.

{\small
	\bibliographystyle{ieee_fullname}
	\bibliography{egbib}
}

\end{document}